\documentclass[10pt,journal]{IEEEtran}
\usepackage{color, soul}
\usepackage{xcolor,cite,etoolbox}
\makeatletter 
\usepackage{cuted}
\usepackage{mathtools}
\newcommand\citecolor[1]{\@namedef{keycolor#1}{\color{black}}}
\makeatother
\usepackage[T1]{fontenc}
\usepackage{textcomp} 
\citecolor{idrissi2023fed}
\citecolor{ahmad2024lightweight}
\citecolor{fallah2020personalized}
\citecolor{wang2021addressing}
\citecolor{djaidja2024federated}
\citecolor{rashid2023federated}
\citecolor{wang2023federated}
\citecolor{wu2021fedkd}
\usepackage{makecell}
\usepackage{threeparttable}
\usepackage{subfigure}
\usepackage{parskip}
\usepackage{svg}
\svgpath{Images/} 
\usepackage{ifpdf}
\usepackage{xcolor}
\usepackage{caption}
\usepackage{subcaption}
\usepackage{amssymb}
\usepackage{pifont}

\usepackage{graphicx}
\usepackage{amsmath}
\usepackage{acronym}
\usepackage{array}
\usepackage{mdwmath}
\usepackage{mdwtab}
\usepackage{xcolor}
\usepackage{multicol}
\usepackage{tabularx}
\usepackage{longtable}
\usepackage{multirow}
\usepackage{subfig}
\usepackage{algorithm}
\usepackage{algorithmic}
\usepackage{etoolbox}
\usepackage{stackengine}
\usepackage{hhline}
\usepackage{booktabs}
\usepackage{longtable}
\usepackage{stackengine}
\usepackage{amssymb}
\usepackage{adjustbox}
\usepackage[font=small,skip=2pt]{caption}
\usepackage[hyphens,spaces,obeyspaces]{url}
\begin{document}
\title{Sentinel: Dynamic Knowledge Distillation for Personalized Federated Intrusion Detection in Heterogeneous IoT Networks}
\author{
Gurpreet Singh,~Keshav Sood,~P. Rajalakshmi,~and~Yong Xiang%
\thanks{Gurpreet Singh, Keshav Sood, and Prof. Yong Xiang are with the School of IT, Deakin University, Australia.
E-mail: zpx@deakin.edu.au, keshav.sood@deakin.edu.au, yong.xiang@deakin.edu.au.
(Corresponding author: Gurpreet Singh.)}%
\thanks{Gurpreet Singh and Prof. P. Rajalakshmi are also with the Indian Institute of Technology Hyderabad, India.
E-mail: id21resch11029@iith.ac.in, raji@ee.iith.ac.in.}
}
\maketitle
\begin{abstract} 
Federated learning (FL) offers a privacy-preserving paradigm for machine learning, but its application in intrusion detection systems (IDS) within IoT networks is challenged by severe class imbalance, non-IID data, and high communication overhead.These challenges severely degrade the performance of conventional FL methods in real-world network traffic classification. To overcome these limitations, we propose Sentinel, a personalized federated IDS (pFed-IDS) framework that incorporates a dual-model architecture on each client, consisting of a personalized teacher and a lightweight shared student model. This design effectively balances deep local adaptation with efficient global model consensus while preserving client privacy by transmitting only the compact student model, thus reducing communication costs. Sentinel integrates three key mechanisms to ensure robust performance: bidirectional knowledge distillation with adaptive temperature scaling, multi-faceted feature alignment, and class-balanced loss functions. Furthermore, the server employs normalized gradient aggregation with equal client weighting to enhance fairness and mitigate client drift. Extensive experiments on the IoTID20 and 5GNIDD benchmark datasets demonstrate that \textit{Sentinel} significantly outperforms state-of-the-art federated methods, establishing a new performance benchmark, especially under extreme data heterogeneity, while maintaining communication efficiency.
\end{abstract}

\IEEEpeerreviewmaketitle
\ifCLASSOPTIONcompsoc
\IEEEraisesectionheading{\section{Introduction}\label{sec:introduction}}
\else

\section{Introduction}
\label{sec:introduction}
\fi
\IEEEPARstart{T}{he} rapid proliferation of billions of heterogeneous Internet of Things (IoT) devices has significantly expanded attack surfaces, presenting new challenges for network security. Insufficient security measures in many of these devices—such as inadequate authentication, weak encryption, and vulnerable communication protocols—facilitate a continuous influx of various cyberattacks, including novel (zero-day) threats, which pose significant risks to the availability, confidentiality, and integrity of data and systems. Traditional firewalls and encryption of the security system are not enough
to prevent increasing cyber attacks \cite{sood2023intrusion}.\par
Intrusion Detection Systems (IDSs) are cost-effective solutions to analyze, monitor traffic patterns, and generate alerts without changing the system topology\cite{10077796}. They are typically deployed on network gateways, routers, servers, and personal or local devices. Often referred to as the first line of defense, IDSs continuously inspect traffic at critical network points to identify malicious or unauthorized activities and potential attacks. 
Traditional machine learning based IDSs require devices to send all local data to a central server for model training. However, privacy concerns and high computational/communication costs make these approaches unsuitable for modern IoT networks. \par
\textcolor{black}{Federated learning based IDSs (Fed-IDSs) have emerged as a promising solution to address these challenges by training a shared global model collaboratively from all clients (IDSs deployed on edge devices) under the orchestration of a central server without transmitting raw data at any central place (server), ensuring data Privacy \cite{kairouz2021advances, 10492991}. 
Despite its potential, Fed-IDSs also face several challenges including statistical heterogeneity due to class imbalance and non-IID data distributions among participating clients, communication overhead due to the large amount of model parameter transfer between clients and server, and susceptibility to data and model poisoning attacks \cite{campos2021evaluating, khraisat2024survey, wang2020attack}}.\par
\textcolor{black}{
Considering statistical heterogeneity, it remains a major challenge in FL. Numerous research efforts have attempted to address this issue, as discussed in the next section.
First, class imbalance occurs when class distributions in local client datasets are non-uniform: one or more classes may contain significantly fewer samples than others (e.g., a long-tail distribution) \cite{shuai2022balancefl}. This imbalance introduces bias from disproportionate class representation and is common in real-world networking data, where malicious samples are far fewer than benign ones. Consequently, the trained model tends to be biased toward the majority class, degrading performance on rare but critical malicious instances \cite{zhang2021fedsens, shuai2022balancefl, tsogbaatar2021iot, 9668482}.
Second, non-IID data arises when clients have distinct (or even unique) data distributions, so local datasets may not reflect the global distribution. The
extreme case of non-IID data is the situation when each client
has unique class data. Challenges in this context emerge from
the imperative to generalize effectively across diverse data
distributions.  These disparities due to class imbalance and non-IID data lead to differences in feature distributions, label distributions, and concept shifts, making it challenging to train a single global model that generalizes well across all clients \cite{li2021ditto, djaidja2024federated, li2022federated, karimireddy2020scaffold}}.\par

\textcolor{black}{To address these challenges, Personalized Federated Learning (pFL) has emerged as a subfield of FL, aiming to mitigate statistical heterogeneity by creating personalized models for each client while still leveraging collaborative learning and preserving data privacy \cite{tan2022towards}. } \textcolor{black}{Existing pFL approaches can be broadly categorized into five main categories, each with its own limitations. For example, \textit{1) Multi-Task and Meta-Learning based approaches} such as FedMTL \cite{smith2017federated} and Per-FedAvg \cite{fallah2020personalized}, aim to learn task relationships across clients or optimize model initialization for rapid adaptation. However, they often assume shared feature representations, which limit their applicability under extreme heterogeneity, and incur high computational costs due to complex optimization processes. \textit{2) Regularization-Based Methods} like Ditto \cite{li2021ditto} and pFedMe \cite{t2020personalized} employ regularization techniques to balance personalization and global model consistency. While effective in certain scenarios, these methods can amplify communication overhead and rely on restrictive assumptions about the convexity of optimization landscapes, which are rarely valid for deep neural networks. \textit{3) Personalized Aggregation approaches} such as FedAMP \cite{huang2021personalized} apply uniform weights across parameters, while FedFomo \cite{zhang2020personalized} and APPLE \cite{luo2022adapt} require transmitting entire models, raising privacy concerns through potential gradient inversion attacks \cite{zhu2019deep}. \textit{4) Model-Splitting strategies} like FedRep \cite{collins2021exploiting} where the idea is to learn a shared “base” or feature representation among clients while allowing each client to personalize only the classifier head. However, in non-IID data, especially when the data is unbalanced, the shared representation can 'drift' because clients learn from very different local distributions. This representation drift can cause the aggregated global model to perform poorly on under‐represented classes \cite{tan2022fedproto}. \textit{5) Knowledge Distillation-based approaches} such as FedKD \cite{wu2022communication} and FedProto \cite{tan2022fedproto} reduce communication costs by exchanging distilled, compact representations rather than full model parameters. However, they can encounter difficulties in aligning knowledge effectively when the featured  spaces across clients are heterogeneous.}\par
Our proposed approach, Sentinel, falls under the knowledge distillation category inspired by FedKD \cite{wu2022communication}. While FedKD introduced a dual-model architecture to reduce communication overhead, its mutual distillation process can amplify prediction ambiguity when teacher-student confidence correlations diverge across clients \cite{chung2025multi}. Moreover, FedKD’s gradient compression, though efficient, may introduce rank deficiency in updates, disproportionately degrading rare-class representations\textcolor{black}{, as demonstrated in our results and analysis section.}

\textcolor{black}{To overcome these challenges, we present Sentinel, a personalized federated learning framework that decouples local adaptation from global knowledge sharing. Each client maintains a personalized teacher model to capture its unique data distribution, while a lightweight student model is shared and aggregated globally to minimize communication cost. At the core of Sentinel lies an adaptive bidirectional knowledge distillation strategy that dynamically regulates distillation intensity based on training progress, inter-model agreement, and prediction confidence. This is further supported by a lightweight feature alignment module designed to bridge representational gaps between heterogeneous teacher and student architecture. Finally, Sentinel integrates a class-balanced loss to address severe data imbalance and applies normalized gradient aggregation at the server, ensuring stable convergence and fairness across highly non-IID client environments.}

In this paper, we present the following key contributions.
\begin{itemize}
    \item \textcolor{black}{\textbf{A Novel pFed-IDS Framework (Sentinel):} We propose an enhanced dual-model architecture for federated intrusion detection that separates local personalization (teacher model) from global aggregation (lightweight student model), significantly reducing communication costs while enhancing performance.}
    
\item \textcolor{black}{\textbf{Multi-objective optimization with adaptive losses:} Sentinel jointly optimizes three complementary objectives. A \textit{class-balanced cross-entropy loss} mitigates traffic imbalance by re-weighting classes for fair treatment of minority attacks. An \textit{adaptive bidirectional knowledge distillation loss} regulates teacher–student knowledge transfer through dynamic temperature scaling and agreement-aware weighting. A \textit{lightweight feature alignment loss} improves representational consistency across heterogeneous models using parameter-efficient mappings with contrastive regularization. Together, these losses form an integrated optimization framework that enhances stability, accuracy, and generalization in non-IID federated IDS settings.  }
    \item \textcolor{black}{\textbf{Fair and Robust Aggregation:} We leverage equal-weight normalized gradient aggregation at the server. By normalizing client updates, this method enhances stability against divergent clients and ensures fairness in heterogeneous environments.}

    \item \textcolor{black}{\textbf{Dynamic Weight Adaptation:} We employ an exponential moving average (EMA) to dynamically adjust the weights for knowledge distillation and feature alignment based on their respective performance scores, allowing the system to self-tune throughout training.}


    \item \textcolor{black}{\textbf{Extensive Empirical Validation:} We conduct extensive experiments on two benchmark networking datasets, demonstrating that Sentinel substantially outperforms eleven state-of-the-art federated learning methods, particularly under challenging non-IID and class-imbalanced conditions.}
\end{itemize} 
\begin{table*}[!th]
    \centering
    \captionsetup{labelfont=bf, font=small}
    \caption{A high-level comparison of our work with existing Federated Learning–based IDSs}
    \label{Summary}
    \resizebox{0.90\textwidth}{!}{%
    \begin{tabular}{
        p{2.3cm} >{\centering\arraybackslash}p{0.8cm} p{2cm} p{2.3cm} p{1.6cm} >{\centering\arraybackslash}p{1cm} >{\centering\arraybackslash}p{2.2cm} >{\centering\arraybackslash}p{1.8cm}>{\centering\arraybackslash}p{1.7cm} >{\centering\arraybackslash}p{1.7cm}
    }\toprule
    \textbf{Fed-IDSs} & \textbf{Year} & \textbf{Target Domain} & \textbf{Dataset} & \textbf{Model Used} & \textbf{Clients} & \textbf{Aggregation method} &\textbf{\textcolor{black}{Address Class Imbalance}}& \textbf{Evaluation on Non-IID data} & \textbf{Communication Efficient} \\\midrule
    \multicolumn{9}{c}{\textbf{Federated Learning–Based IDSs}} \\\midrule
    Rahman et al. \cite{rahman2020internet}  & 2020 & IoT Networks & NSL KDD & DNN & 4 & FedAvg & \ding{55}& \ding{51} & \ding{55} \\\midrule
    Li et al. \cite{li2020deepfed}         & 2020 & Industrial Cyber Physical Systems & Gas Pipeline & CNN, GRU and MLP & 3, 5, and 7 & - & \ding{55}& \ding{55} & \ding{55} \\\midrule
    Li et al. \cite{li2021fleam}                & 2021 & Industrial IoT & UNSW-NB15 & GRU & 4 & - & \ding{55}& \ding{55} & \ding{55} \\\hline
    Campos et al. \cite{campos2021evaluating}   & 2021 & IoT Networks & CIC-ToN IoT & Logistic Regression & 10 & Fed+ & \ding{55}& \ding{51} & \ding{55} \\\midrule
    Popoola et al. \cite{9499122}               & 2022 & IoT Networks & BoT-IoT and N-BaIoT & DNN & 5 & FedAvg & \ding{55}& \ding{55} & \ding{55} \\\midrule
    Rey et al. \cite{rey2022federated}          & 2022 & IoT Networks & N-BaIoT & MLP, Autoencoder & 9 & FedAvg & \ding{55}& \ding{55} & \ding{55} \\\midrule
    Friha et al. \cite{friha2022felids}         & 2022 & Agriculture IoT Networks & CICIDS 2018, MQTT, InSDN & DNN, CNN, RNN & 5, 10, and 15 & FedAvg& \ding{55} & \ding{51} & \ding{55} \\\midrule
    Liu et al. \cite{liu2022intrusion}          & 2022 & Maritime Transportation & NSL KDD & CNN, MLP & 100 & FedBatch & \ding{55}& \ding{51} & \ding{55} \\\midrule
    Idrissi et al. \cite{idrissi2023fed}         & 2023 & Computer Networks & USTC-TFC2016, CIC-IDS17/18 & Autoencoder & 10 & FedAvg, FedProx & \ding{55}& \ding{55} & \ding{55} \\\midrule
    Rashid et al. \cite{rashid2023federated}     & 2023 & IoT Networks & Edge IIoTset & CNN, RNN & 3, 9, and 15 & FedAvg & \ding{55}& \ding{51} & \ding{55} \\\midrule
    Ahmad et al. \cite{ahmad2024lightweight}     & 2024 & IoT Networks & ToN-IoT, N-BaIoT & MLP & 3 & FedAvg & \ding{55}& \ding{55} & \ding{55} \\\midrule
    Djaidja et al. \cite{djaidja2024federated}   & 2024 & 5G and Beyond & NSL KDD & MLP & 8 & FedAvg & \ding{55}& \ding{51} & \ding{55} \\\midrule
    Singh et al. \cite{10492991}                 & 2024 & IoT Networks & IoTID20, WUSTL 2021, Edge-IIoTset, 5GNIDD & DNN & 15 & FedAvg & \ding{51}& \ding{51} & \ding{55} \\\midrule
    \multicolumn{9}{c}{\textbf{Personalized Federated Learning–Based IDSs}} \\\midrule 
    Shibly et al. \cite{10056726}                & 2022 & Automotive & Own dataset & CNN, MLP, XGBoost, AE & 3 & - & \ding{55}& \ding{55} & \ding{55} \\\midrule
   \textcolor{black}{ Huang et al. \cite{9919869}  }                & \textcolor{black}{2022} & \textcolor{black}{Industrial Cyber Physical Systems} & \textcolor{black}{NSLKDD, SWAT, and WADI}& \textcolor{black}{CNN} & \textcolor{black}{5} & - & \ding{55}& \textcolor{black}{\ding{51} }& \textcolor{black}{\ding{55}} \\\midrule
    Hoang et al. \cite{hoang2023canperfl}        & 2023 & Automotive & Own dataset & CNN & 3 & FedAvg & \ding{55}& - & \ding{55} \\\midrule
    Singh et al. \cite{10469061}& 2023 & Metaverse & Edge-IIoTset, 5GNIDD& DNN& 20 & FedAvg & \ding{55}&\ding{51}&  \ding{51} \\\midrule
    \textcolor{black}{Yan et al. \cite{yan2024global}} & \textcolor{black}{2024}& \textcolor{black}{IoT Networks} & \textcolor{black}{Bot-IoT, UNSW-NB15, CICIDS2017}& \textcolor{black}{NN}& \textcolor{black}{50,100, 200,500}& \textcolor{black}{FedAvg}& \ding{55}&\textcolor{black}{ \ding{51}}& \textcolor{black}{\ding{55}} \\\midrule
    \textcolor{black}{Thein et al. \cite{thein2024personalized}} & \textcolor{black}{2024}& \textcolor{black}{Computer Networks} & \textcolor{black}{NBaIoT}& \textcolor{black}{1DCNN}& \textcolor{black}{20}& \textcolor{black}{FedAvg with poisoined client detector}& \ding{55}&\textcolor{black}{ \ding{51}}& \textcolor{black}{\ding{55}} \\\midrule
    Sentinel (Ours)                             & -    & Heterogeneous IoT Networks & IoT-ID-20, 5GNIDD & DNNs, CNN-MLP & 10 & FedNGA \cite{zuo2024byzantine} (Byzantine Robust)& \ding{51} & \ding{51} & \ding{51} \\\bottomrule  
    \end{tabular}}
\end{table*}
The remainder of the paper is structured as follows: Section II discusses related work. Table \ref{Summary} presents a high-level comparison of our approach with existing methods, while Table \ref{tab:notations} defines the key mathematical notations. Section III presents the \textit{Sentinel} framework along with its mathematical foundations. Section IV reports the experimental results, including implementation details, ablation studies, and complexity analysis. Finally, Section V concludes the paper.
\section{Related Work}
In recent years, Federated Learning has gained a significant attention for enhancing security and privacy in distributed IoT networks. 
Several surveys \cite{khraisat2024survey, belenguer2025review} reviews the latest advancement in this domain. 
\subsection{Federated learning based IDSs}
Rey et al. \cite{rey2022federated} investigated the effectiveness of Fed-IDS for detecting malware attacks on IoT devices, comparing traditional centralized, distributed, and federated approaches using the N-BaIoT dataset \cite{meidan2018n}. They assessed model resilience against adversarial attacks and explored various aggregation techniques, but did not consider non-IID data distributions.
Friha et al. \cite{friha2022felids} introduced a Fed-IDS named FELIDS for agricultural IoT networks, demonstrating superior performance over conventional ML/DL-based IDSs. Their study evaluated three deep learning models with FedAvg across three datasets, but relied on oversampling to address class imbalance.
The authors in \cite{9499122} demonstrated FL's effectiveness in detecting zero-day botnet attacks using DNNs. They created a zero-day attack scenario by excluding one class from each client during data partitioning. However, the details about selecting class samples from the nine clients of the NBaIoT dataset are not provided. 

Rashid et al. \cite{rashid2023federated} proposed a Fed-IDS for IIoT networks, employing FedAvg to aggregate model parameters from edge devices.
Their study demonstrated that the federated approach achieved comparable accuracy to centralized ML models while enhancing data privacy and reducing bandwidth requirements. Aouedi et al. \cite{aouedi2022federated} implemented a semi-supervised FL framework for intrusion detection in industrial IoT systems, leveraging unsupervised autoencoder (AE) models at edge devices to learn latent representations of network traffic patterns. Their approach utilized federated aggregation of locally trained AE parameters to construct a global feature extractor, which was then combined with a small centralized labeled dataset at the server for supervised classification. However, performance degradation may occur under strong non-IID data distributions due to inconsistent feature space alignment across devices.


\subsection{Personalized Federated Learning-based IDSs}  
Thein et al. \cite{thein2024personalized} propose pFed-IDS to tackle the challenges of data heterogeneity and poisoning attacks in IoT networks. Their method leverages a logit adjustment loss to customize local models to client-specific non-IID data distributions, thereby enhancing personalization. Additionally, they introduce a server-side poisoned client detector that employs a two-step similarity alignment process: first, identifying potentially clean clients by analyzing their proximity to a pre-computed global model, and second, computing angular deviations from the centroid of non-poisoned client updates to re-weight contributions during aggregation. 
However, the approach assumes that poisoned client updates exhibit detectable angular deviations — a vulnerability that advanced adversarial strategies could exploit and incurs increased computational overhead due to pairwise similarity calculations, which may be challenging in large-scale IoT deployments. Huang et al. \cite{9919869} present EEFED, a pFL framework for CPS intrusion detection, addressing challenges posed by non-IID data distributions. The framework introduces an Execution and Evaluation dual network, where the execution network facilitates global model updates while preserving privacy, and the evaluation network dynamically optimizes local models using a reinforcement learning-inspired backtracking replacement algorithm. Additionally, a personalized update algorithm adjusts models based on environment similarity parameters to enhance local adaptation. 
While EEFED effectively enhances detection accuracy and model stability, potential limitations include increased computational complexity due to dual-network synchronization and reliance on accurate similarity parameter estimation for optimal performance in highly heterogeneous environments.
Yan et al. \cite{yan2024global} propose a client-sampled federated meta-learning framework for personalized IoT intrusion detection, balancing global knowledge sharing with local adaptation. Their approach mitigates Non-IID data issues through a two-stage optimization: selecting representative clients via adaptive sampling based on data diversity and computational capacity, followed by model-agnostic meta-learning (MAML) \cite{finn2017model} for rapid adaptation to device-specific attack patterns.  
However, the computational complexity of meta-learning and clustering mechanism introduces significant overhead for resource-constrained IoT devices during local updates.
 
 \par
 \begin{figure*}[h!]
\centering
\includegraphics[width=4.8in]{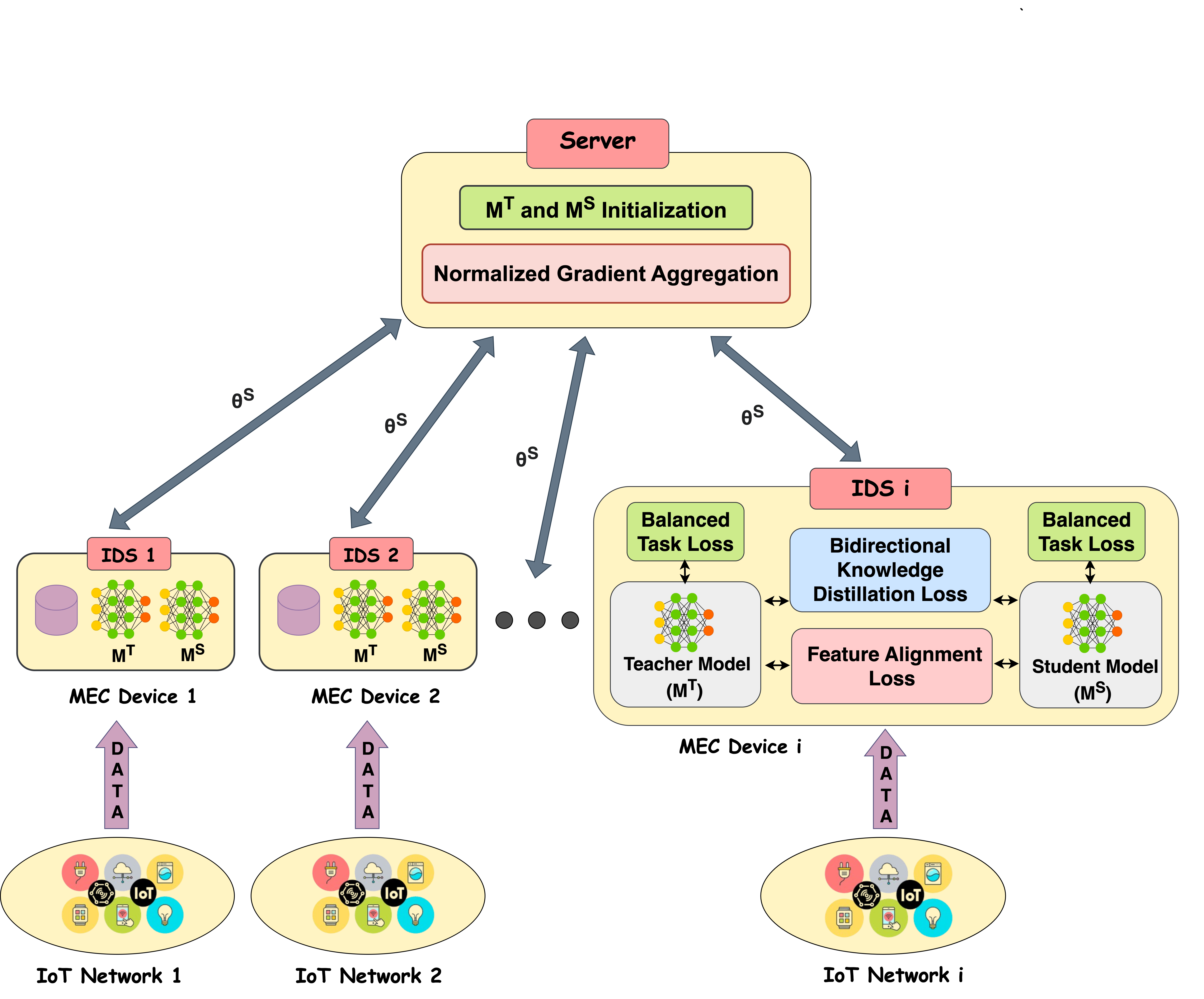}
\caption{High-Level architecture and workflow of the Sentinel.}
\label{5Garc}
\end{figure*} 

\section{Design of Sentinel}

\subsection{System Overview}
Fig. \ref{5Garc} illustrates the high-level deployment framework of Sentinel, which employs a three-layer architecture for intrusion detection in heterogeneous IoT networks. The lowest layer represents the IoT devices, encompassing a diverse range of intelligent endpoints within each IoT network. The middle layer consists of MEC (Mobile Edge Computing) devices or platforms that host the IDS, which can be referred to as a client in the context of federated learning (FL). At the topmost layer lies a security cloud platform (Server), managed by the network operator, to oversee and coordinate the overall system.
In this framework, each MEC platform collects traffic data (PCAP files) from its associated IoT networks, converts it into CSV format, pre-processes it, and trains a local IDS with both teacher and student models. Owing to their computational resources, MEC devices can efficiently store and process data from their respective IoT networks.\par
\begin{table}[t]
\centering
\footnotesize
\caption{\textcolor{black}{Key Mathematical Notations}}
\label{tab:notations}
\resizebox{0.48\textwidth}{!}{%
\begin{tabular}{@{} l p{6.8cm} @{}}
\toprule
\textbf{Symbol} & \textbf{Explanation} \\
\midrule
\multicolumn{2}{l}{\textit{\textbf{Models, Data, and Features}}} \\
$M_T, M_S$ & Teacher model (personalized) and student model (shared). \\
$\theta_T, \theta_S$ & Parameters of the teacher and student models. \\
$h_T, h_S$ & Feature representations from model backbones. \\
$z_T, z_S$ & Logits (pre-softmax outputs) from model heads. \\
$\tilde{h}_T, \tilde{h}_S$ & $\ell_2$-normalized features for contrastive learning. \\
$\mathcal{D}_i$ & Local dataset for client $i$. \\
$n_c$ & Number of samples for class $c$ on a local client. \\
$\mathcal{M}_T, \mathcal{M}_S$ & Memory banks storing past teacher and student features. \\
\midrule
\multicolumn{2}{l}{\textit{\textbf{Loss Components and Hyperparameters}}} \\
$\mathcal{L}_{\text{task}}$ & Class-balanced cross-entropy task loss. \\
$\mathcal{L}_{\text{KD}}$ & Adaptive bidirectional knowledge distillation loss. \\
$\mathcal{L}_{\text{align}}$ & Multi-component feature alignment loss. \\
$w_c, \beta$ & Class weight for class $c$ and its rebalancing hyperparameter (0.999). \\
$T_r$ & KD temperature at round $r$ (decays from $T_{\text{base}}=3.0$ to $T_{\min}=1.0$). \\
$\tau$ & Temperature for contrastive loss (fixed at 0.1). \\
$\gamma$ & Teacher-student prediction agreement rate. \\
$\delta$ & Mean product of max probabilities (joint confidence). \\
$\lambda_{\text{KD}}, \lambda_{\text{align}}$ & Dynamically adapted weights for KD and alignment losses. \\
$\alpha_r$ & EMA smoothing factor at round $r$ (decays from $0.9 \to 0.7$). \\
$f_{\text{align}}$ & Lightweight feature alignment module. \\
$D_{\text{KL}}(\cdot \| \cdot)$ & Kullback-Leibler divergence. \\
\midrule
\multicolumn{2}{l}{\textit{\textbf{Server-Side Aggregation}}} \\
$g_i$ & Client pseudo-gradient, defined as $\theta_S^{\text{global}} - \theta_S^{i}$. \\
$\hat{g}_i$ & $\ell_2$-normalized client pseudo-gradient. \\
$\bar{g}$ & Aggregated normalized pseudo-gradient at the server. \\
$v^{(r)}$ & Server-side momentum buffer at round $r$. \\
$\eta$ & Server-side learning rate (step size). \\
\midrule
\multicolumn{2}{l}{\textit{\textbf{Federated Learning Parameters}}} \\
$r, E$ & Current communication round and local training epochs. \\
$\mathcal{S}_r, \mathcal{U}_r$ & Sets of selected clients and reliable clients (after filtering). \\
$\rho, p_{\text{drop}}$ & Client participation ratio and dropout probability. \\
$\epsilon$ & Small constant for numerical stability ($10^{-8}$). \\
\bottomrule
\end{tabular}
}
\end{table}

The extensive communication between the server and clients required to exchange model updates becomes a significant bottleneck, particularly for large-scale models with billions of parameters. This imposes a considerable computational and bandwidth burden on resource-constrained clients. To address this challenge, Sentinel employs a dual-model architecture, where only the lightweight student model is shared with the central server during federated learning. The student model is designed to have a similar architecture to the teacher model but with reduced capacity (e.g., fewer layers or neurons), thereby minimizing communication overhead while maintaining compatibility for knowledge distillation. The teacher model, which is more complex and personalized for local training, remains on the client device and is not shared with the server. This approach ensures efficient collaboration between clients and the server while reducing resource consumption, making Sentinel particularly well-suited for federated learning in resource-constrained environments.

\subsection{Client-Side Training in Sentinel: Theory and Design}

In Sentinel, each client device (e.g., an edge device in an IoT network)
maintains two neural network models:
a personalized \emph{teacher model} ($M_T$) trained primarily on the
client’s local data, and a \emph{student model} ($M_S$) that participates
in global knowledge sharing across all clients.
The design goal is to simultaneously maximize
\emph{local performance} (personalization) and
\emph{global detection capability} (collaboration)
through a joint optimization process.

To achieve this, Sentinel defines three complementary objective functions
(losses) that operate together during client-side training.

\subsubsection{\textcolor{black}{Class-Balanced Task Loss for Data Imbalance}}
\textcolor{black}{Network traffic datasets are inherently imbalanced, with benign samples often vastly outnumbering malicious ones~\cite{10492991}. To prevent models from developing a bias toward the majority class, \textit{Sentinel} employs a class-balanced cross-entropy (CBCE) loss based on the concept of an ``effective number of samples''~\cite{cui2019class}. For each class $c$ with $n_c$ local samples, a weight $w_c$ is calculated as:
\begin{equation}
w_c = \frac{1 - \beta}{1 - \beta^{n_c}},
\end{equation}
where $\beta \in [0,1)$ is a hyperparameter controlling the re-weighting intensity. We set $\beta=0.999$, a value close to 1, to strongly up-weight minority classes with few samples. To prevent gradient instability from extreme weights, the computed weights are clamped to $[0.2, 5.0]$ and subsequently mean-normalized, which preserves the overall scale of the standard cross-entropy loss.}

\textcolor{black}{The final task loss, $\mathcal{L}_{\text{task}}$, is the average of the class-balanced losses for both the teacher and student models, ensuring both are guided by the same balanced objective:
\begin{equation}
\mathcal{L}_{\text{task}}
= \frac{1}{2} \left[
\mathcal{L}_{\text{CBCE}}(\mathbf{z}_T, y; \mathbf{w})
+ \mathcal{L}_{\text{CBCE}}(\mathbf{z}_S, y; \mathbf{w})
\right],
\end{equation}
where $\mathbf{z}_T$ and $\mathbf{z}_S$ are the logits, $y$ is the ground-truth label, and $\mathbf{w} = \{w_c\}_{c=1}^C$ is the vector of class weights.}

\subsubsection{\textcolor{black}{Adaptive Bidirectional Knowledge Distillation}}
\textcolor{black}{To facilitate knowledge transfer between the local teacher and the global student, \textit{Sentinel} introduces an adaptive, bidirectional distillation mechanism, governed by two key components.}

\paragraph{\textcolor{black}{Temperature Scaling with Decay.}}
\textcolor{black}{Following standard KD practices~\cite{hinton2015distilling}, a temperature parameter $T$ is used to soften the model outputs, encouraging the student to learn from the teacher's class distributions. The temperature is dynamically adjusted across communication rounds $r$ to transition from exploring general patterns (high $T$) to refining specific knowledge (low $T$):
\begin{equation}
T_r = \max \!\left( T_{\min}, \; T_{\text{base}} \cdot T_{\text{decay}}^{\,\min\!\left(\tfrac{r}{10}, 5\right)} \right)
\end{equation}
The parameters are chosen for a smooth annealing process: a base temperature $T_{\text{base}}=3.0$ and minimum $T_{\min}=1.0$ balance initial exploration with final performance. The decay factor $T_{\text{decay}}=0.95$ ensures a gentle decay. The exponent's scaling ($r/10$) and cap (5) concentrate the annealing over the first 50 rounds and prevent overly aggressive decay in later stages, thereby promoting stability.}

\textcolor{black}{\paragraph{Confidence-Weighted Bidirectional KD Loss.}
\textit{Sentinel}'s distillation loss adapts to the models' real-time performance. Let $p_T$ and $p_S$ be the temperature-scaled probability outputs. We define two metrics: agreement ($\gamma$), the fraction of samples where both models predict the same class, and joint confidence ($\delta$), the geometric mean of their confidence on the predicted classes. The bidirectional loss is then:
\begin{equation}
\mathcal{L}_{\text{KD}} = \bar{w}_y \cdot \Big[
(1-\gamma) \cdot D_{\text{KL}}(p_S \parallel p_T)
+ \delta \cdot D_{\text{KL}}(p_T \parallel p_S)
\Big] \cdot T^2
\end{equation}
where $D_{\text{KL}}$ is the Kullback–Leibler divergence and $\bar{w}_y$ is the mean class weight for the samples in the current batch. This formulation has an intuitive effect: The weight $(1-\gamma)$ for teacher-to-student distillation ensures that the teacher's guidance is strongest when the student disagrees, focusing learning on correcting mistakes. Conversely, the weight $\delta$ for student-to-teacher distillation ensures the student only ``teaches back'' when both models are highly confident, preventing noisy signals from harming the specialized teacher. The loss is modulated by the mean class weight $\bar{w}_y$ to counteract class imbalance and by $T^2$ to correctly scale gradients.}
\subsubsection{\textcolor{black}{Multi-Component Feature Alignment}}
\textcolor{black}{Since the teacher ($M_T$) and student ($M_S$) models can have heterogeneous architectures, their intermediate features ($h_T, h_S$) are not directly comparable. To bridge this gap, \textit{Sentinel} enforces alignment in their embedding spaces using a multi-component loss designed to capture geometric, directional, and structural similarity:
\begin{equation}
\mathcal{L}_{\text{align}} = \mathcal{L}_{\text{geom}} + \lambda_{\text{cos}} \mathcal{L}_{\text{dir}} + \lambda_{\text{contrast}} \mathcal{L}_{\text{struct}},
\end{equation}
where $\lambda_{\text{cos}}=0.5$ and $\lambda_{\text{contrast}}=0.2$. Each component serves a distinct purpose in a hierarchical alignment strategy:}

\textcolor{black}{\paragraph{Geometric Alignment ($\mathcal{L}_{\text{geom}}$).}
The primary goal is to enforce strict geometric proximity. A parameter-efficient \textit{LightweightAligner} first maps teacher features to the student's space, $h'_T = f_{\text{align}}(h_T)$. For each projected teacher feature $h'_{T,i}$ in a batch, we dynamically find its nearest-neighbor student feature, $h^*_{S,i}$, a technique inspired by FedDNA~\cite{wang2023feddna}. The geometric loss then minimizes the Mean Squared Error between these matched pairs:
\begin{equation}
    \mathcal{L}_{\text{geom}} = \frac{1}{B}\sum_{i=1}^B \|h'_{T,i} - h^*_{S,i}\|_2^2.
\end{equation}}

\textcolor{black}{\paragraph{Directional Alignment ($\mathcal{L}_{\text{dir}}$).}
As a secondary, less stringent regularizer, we enforce directional consistency. This relaxes the magnitude constraint of MSE, encouraging the matched feature vectors ($h'_{T,i}, h^*_{S,i}$) to point in the same direction:
\begin{equation}
    \mathcal{L}_{\text{dir}} = 1 - \frac{1}{B}\sum_{i=1}^B \frac{h'_{T,i} \cdot h^*_{S,i}}{\|h'_{T,i}\|_2 \, \|h^*_{S,i}\|_2}.
\end{equation}}

\textcolor{black}{\paragraph{Structural Alignment ($\mathcal{L}_{\text{struct}}$).}
Finally, a memory-augmented contrastive loss refines the fine-grained structure of the embedding space. For each sample $i$, this loss attracts the feature pair $(h'_{T,i}, h_{S,i})$ derived from the same input, while repelling it from other in-batch features and a large set of negative features stored in a memory bank $\mathcal{M}_S$. We use the InfoNCE loss~\cite{oord2018representation}, which is expressed as:
\begin{equation}
\mathcal{L}_{\text{struct}}
= -\frac{1}{B}\sum_{i=1}^B 
\log \frac{\exp(\tilde{h}'_{T,i}\cdot \tilde{h}_{S,i}/\tau)}
{P_i + N_i},
\end{equation}
where the denominator terms are the sums of exponentiated similarities to all in-batch student features ($P_i$) and all memory bank features ($N_i$):
\begin{equation}
P_i = \sum_{j=1}^B \exp(\tilde{h}'_{T,i}\cdot \tilde{h}_{S,j}/\tau),
\quad
N_i = \sum_{\mathbf{m}\in\mathcal{M}_S} \exp(\tilde{h}'_{T,i}\cdot \tilde{\mathbf{m}}/\tau).
\end{equation}
Here, $\tilde{h}$ denotes an $\ell_2$-normalized feature and $\tau=0.1$ is a temperature parameter that sharpens the similarity distribution to improve feature discrimination.}

\begin{algorithm}[t!]
\caption{Sentinel: Client-Side Training}
\label{alg:client_training}
\begin{algorithmic}[1]
\footnotesize
\REQUIRE Local dataset $\mathcal{D}_i$, local epochs $E$, teacher model $M_T$, student model $M_S$, aligner $f_{\text{align}}$
\REQUIRE Global round $r$, memory banks $\mathcal{M}_T$, $\mathcal{M}_S$
\STATE \textbf{Initialize:} $\lambda_{\text{KD}} \gets 0.2$, $\lambda_{\text{align}} \gets 0.08$, $T_{\text{base}} \gets 3.0$, $T_{\min} \gets 1.0$, $T_{\text{decay}} \gets 0.95$
\STATE \textbf{Compute class weights} $\mathbf{w} = \{w_c\}$ using effective sample rebalancing with $\beta = 0.999$
\FOR{epoch $e = 1$ to $E$}
  \FOR{each mini-batch $(X, Y) \in \mathcal{D}_i$}
    
    \STATE \textcolor{black}{\textbf{// Forward Pass \& Feature Extraction}}
    \STATE Extract features: $h_T \gets M_T^{\text{base}}(X)$, $h_S \gets M_S^{\text{base}}(X)$
    \STATE Generate predictions: $z_T \gets M_T^{\text{head}}(h_T)$, $z_S \gets M_S^{\text{head}}(h_S)$
    
    \STATE \textcolor{black}{\textbf{// 1. Task Loss (Class-Balanced)}}
    \STATE $\mathcal{L}_{\text{task}} \gets \frac{1}{2}\left[\mathcal{L}_{\text{CE}}(z_T, Y; \mathbf{w}) + \mathcal{L}_{\text{CE}}(z_S, Y; \mathbf{w})\right]$
    
    \STATE \textcolor{black}{\textbf{// 2. Adaptive Knowledge Distillation}}
    \STATE Update temperature: $T_r \gets \max(T_{\min}, T_{\text{base}} \cdot T_{\text{decay}}^{\min(r/10, 5)})$
    \STATE Compute agreement $\gamma$ and confidence $\delta$ from $z_T, z_S$
    \STATE $\mathcal{L}_{\text{KD}} \gets$ Bidirectional KL-divergence weighted by $\gamma, \delta$, and $\bar{w}_y$
    
    \STATE \textcolor{black}{\textbf{// 3. Feature Alignment}}
    \STATE Project teacher features: $h'_T \gets f_{\text{align}}(h_T.\text{detach()})$ 
    \STATE Find NN-matched student features $h^*_S$ for $h'_T$ and compute alignment score $\text{score}_{\text{align}}$
    \STATE $\mathcal{L}_{\text{align}} \gets \mathcal{L}_{\text{MSE}}(h'_T, h^*_S) + \lambda_{\text{cos}}\mathcal{L}_{\text{cos}}(h'_T, h^*_S) + \lambda_{\text{contrast}}\mathcal{L}_{\text{contrastive}}(h'_T, h_S, \mathcal{M})$
    \STATE Update memory banks: $\mathcal{M}_T \gets \mathcal{M}_T \cup h_T$, $\mathcal{M}_S \gets \mathcal{M}_S \cup h_S$
    
    \STATE \textcolor{black}{\textbf{// 4. Dynamic Weight Adaptation}}
    \STATE Adapt weights using EMA: $\lambda_{\text{KD}} \gets \alpha_r\lambda_{\text{KD}} + (1-\alpha_r)(1-\gamma)$
    \STATE \phantom{Adapt weights using EMA: }$\lambda_{\text{align}} \gets \alpha_r\lambda_{\text{align}} + (1-\alpha_r)(1-\text{score}_{\text{align}})$
    \STATE Clamp weights to dynamic bounds: $\lambda_{\text{KD}} \in [\lambda_{\text{min}}, \lambda_{\text{max}}(r)]$, $\lambda_{\text{align}} \in [\lambda_{\text{min}}, \lambda_{\text{max}}(r)]$
    
    \STATE \textcolor{black}{\textbf{// 5. Combined Loss \& Optimization}}
    \STATE $\mathcal{L}_{\text{total}} \gets \mathcal{L}_{\text{task}} + \lambda_{\text{KD}} \cdot \mathcal{L}_{\text{KD}} + \lambda_{\text{align}} \cdot \mathcal{L}_{\text{align}}$
    \STATE Update parameters of $M_T, M_S, f_{\text{align}}$ via AdamW with gradient clipping
    
  \ENDFOR
  \STATE Update learning rate scheduler
\ENDFOR
\RETURN Updated student model state dictionary $\text{state\_dict}(M_S)$
\end{algorithmic}
\end{algorithm}
\subsubsection{\textcolor{black}{Adaptive Loss Balancing and Final Objective}}
\textcolor{black}{The full client optimization objective in \textit{Sentinel} integrates the task, distillation, and alignment losses into a single, cohesive framework:
\begin{equation}
\mathcal{L}_{\text{total}} = \mathcal{L}_{\text{task}} + \lambda_{\text{KD}} \mathcal{L}_{\text{KD}} + \lambda_{\text{align}} \mathcal{L}_{\text{align}}.
\end{equation}
A key innovation is that the auxiliary loss weights, $\lambda_{\text{KD}}$ and $\lambda_{\text{align}}$, are not fixed. Instead, they evolve dynamically based on real-time model performance, updated each batch via an Exponential Moving Average (EMA):
\begin{align}
\lambda_{\text{KD}} &\leftarrow \alpha_r \lambda_{\text{KD}} + (1-\alpha_r)(1-\gamma), \\
\lambda_{\text{align}} &\leftarrow \alpha_r \lambda_{\text{align}} + (1-\alpha_r)(1-\text{score}_{\text{align}}).
\end{align}
This creates a self-regulating feedback loop: if a component underperforms (e.g., low teacher-student agreement $\gamma$), its target weight increases, automatically amplifying the optimization pressure. }

\textcolor{black}{To ensure this dynamic process remains stable, it is governed by two carefully designed safeguards:}
\textcolor{black}{\paragraph{Dynamic EMA Smoothing.} The smoothing factor, $\alpha_r = \max(0.7, 0.9 - 0.03r)$, starts high for stability in early rounds when models are noisy, and decreases to allow more responsive updates as training progresses.
\paragraph{Dynamic Range Control.} The weights are clamped within dynamically expanding bounds that enforce both a persistent learning signal (non-zero lower bound) and prevent auxiliary losses from overpowering the primary task (adaptive upper bound):
\begin{align}
\lambda_{\text{KD}} &\in [0.03, \min(0.35, 0.18 + 0.02r)], \\
\lambda_{\text{align}} &\in [0.01, \min(0.12, 0.06 + 0.01r)].
\end{align}
This principled adaptive formulation enables \textit{Sentinel} to maintain a strong balance between local personalization, collaborative knowledge sharing, and representation consistency, making it resilient to the challenges of heterogeneity and class imbalance.}

\subsection{\textcolor{black}{Server-Side Robust Aggregation in Sentinel}}
\textcolor{black}{The server in \textsc{Sentinel} coordinates collaborative learning while remaining resilient to client failures, communication bottlenecks, and adversarial updates. The aggregation procedure is designed specifically for the unique challenges of federated intrusion detection systems (IDS).}

\textcolor{black}{\paragraph{Client Selection and Model Distribution.}
At the beginning of each round $r$, the server selects a subset of clients $\mathcal{S}_r$ according to a predefined join ratio $\rho$. Only the lightweight \emph{student model parameters} $\theta_S^{\text{global}}$ are distributed to these clients. This choice significantly reduces communication overhead, as the larger, personalized teacher models remain local to each client.}

\textcolor{black}{\paragraph{Robust Client Filtering.}
Once clients complete their local training and send updates, the server constructs a reliable client set $\mathcal{U}_r$ by filtering out dropouts and stragglers. From the initial selection $\mathcal{S}_r$, an active set $\mathcal{A}_r$ is first formed by accounting for dropouts with probability $p_{\text{drop}}$. Stragglers are then removed based on their average per-round execution time $t_c$. The final set of clients for aggregation is thus defined as:
\begin{equation}
    \mathcal{U}_r = \{c \in \mathcal{A}_r \mid t_c \le T_{\text{thresh}}\}.
\end{equation}
In our experiments, we set $p_{\text{drop}}=0$ and a permissive $T_{\text{thresh}}=10,000$s. This effectively deactivates filtering within our controlled 10-client environment, allowing the evaluation to focus on the performance of the core distillation and aggregation framework.}

\begin{algorithm}[t!]
\caption{Server-Side Aggregation in Sentinel}
\label{alg:server_aggregation}
\begin{algorithmic}[1]
\footnotesize
\REQUIRE Total clients $\mathcal{C}$, global rounds $R$, join ratio $\rho$, server LR $\eta$, momentum $\beta$
\REQUIRE Global student model $M_S^{\text{global}}$, time threshold $T_{\text{thresh}}$, client drop rate $p_{\text{drop}}$
\STATE \textbf{Initialize:} Server momentum buffer $v^{(0)} \gets \mathbf{0}$

\FOR{round $r = 1$ to $R$}
    \STATE \textcolor{black}{\textbf{// 1. Client Selection \& Model Distribution}}
    \STATE Select participating clients: $\mathcal{S}_r \gets \textsc{SelectClients}(\mathcal{C}, \rho)$
    \STATE Broadcast global student model parameters $\theta_S^{\text{global}}$ to all clients in $\mathcal{S}_r$
    
    \STATE \textcolor{black}{\textbf{// 2. Parallel Client Training}}
    \STATE \textbf{for all} clients $i \in \mathcal{S}_r$ \textbf{in parallel do}
        \STATE $\theta_S^i \gets \textsc{ClientUpdate}(i, \theta_S^{\text{global}}, r)$ \textcolor{gray}{// Run Algorithm 1}
    \STATE \textbf{end for}
    
    \STATE \textcolor{black}{\textbf{// 3. Robust Model Reception \& Filtering}}
    \STATE Simulate client dropouts: $\mathcal{A}_r \gets \textsc{RandomSample}(\mathcal{S}_r, 1-p_{\text{drop}})$
    \STATE Filter by time constraint: $\mathcal{U}_r \gets \{i \in \mathcal{A}_r : \text{TrainingTime}(i) \leq T_{\text{thresh}}\}$
    \STATE Collect updated models $\{\theta_S^i : i \in \mathcal{U}_r\}$
    
    \STATE \textcolor{black}{\textbf{// 4. Normalized Gradient Aggregation}}
    \STATE Compute client pseudo-gradients: $g_i \gets \theta_S^{\text{global}} - \theta_S^i, \; \forall i \in \mathcal{U}_r$
    \STATE Normalize gradients: $\hat{g}_i \gets g_i / (\|g_i\|_2 + \epsilon), \; \forall i \in \mathcal{U}_r$
    \STATE Aggregate with equal weight: $\bar{g} \gets \frac{1}{|\mathcal{U}_r|} \sum_{i \in \mathcal{U}_r} \hat{g}_i$
    
    \STATE \textcolor{black}{\textbf{// 5. Momentum-Enhanced Update}}
    \STATE Update momentum buffer: $v^{(r)} \gets \beta \cdot v^{(r-1)} + (1-\beta) \cdot \bar{g}$
    \STATE Update global model: $\theta_S^{\text{global}} \gets \theta_S^{\text{global}} - \eta \cdot v^{(r)}$
    
\ENDFOR
\RETURN Trained global student model parameters $\theta_S^{\text{global}}$
\end{algorithmic}
\end{algorithm}
\textcolor{black}{\paragraph{Normalized Gradient Aggregation.}
Instead of standard FedAvg, which can be skewed by client heterogeneity, \textsc{Sentinel} employs \emph{normalized gradient aggregation} on the models from reliable clients ($\mathcal{U}_r$). First, each local update is computed as a pseudo-gradient: $g_i = \theta_S^{\text{global}} - \theta_S^i$. To prevent any single client from dominating the update, each gradient is then scaled to have a unit $\ell_2$-norm:
\begin{equation}
    \hat{g}_i = \frac{g_i}{\lVert g_i \rVert_2 + \epsilon},
\end{equation}
where $\epsilon$ is a small constant for numerical stability. The server aggregates these normalized gradients with equal weight:
\begin{equation}
    \bar{g} = \frac{1}{|\mathcal{U}_r|} \sum_{i \in \mathcal{U}_r} \hat{g}_i .
\end{equation}
This design provides robustness against differing local data distributions and client-side optimizers, and offers resilience to adversarial (Byzantine) updates — malicious or faulty clients updates that behave arbitrarily and may attempt to poison the global model by submitting misleading or abnormally large gradients.}

\textcolor{black}{\paragraph{Momentum-Enhanced Convergence.}
To stabilize the optimization trajectory and accelerate convergence, the server maintains a momentum buffer $v$, which is updated using the aggregated gradient:
\begin{equation}
    v^{(r)} = \beta \cdot v^{(r-1)} + (1-\beta)\cdot \bar{g},
\end{equation}
where $v^{(0)}$ is initialized to zero and $\beta=0.9$ is a standard momentum value. The global model is then updated using this smoothed direction:
\begin{equation}
    \theta_S^{\text{global}} \leftarrow \theta_S^{\text{global}} - \eta \cdot v^{(r)}.
\end{equation}
The server learning rate $\eta$ acts as a global step size. We set $\eta=1.0$, a common choice for normalized aggregation schemes, as it applies a consistent step in the aggregated direction determined by the clients.}

\section{Experimental Results and Analysis}  
Experiments were conducted on a Linux server (Ubuntu 22.04.2 LTS) with dual Intel Xeon Platinum 8168 CPUs and 1.5~TB DDR4 memory. Data normalization was performed using Standard Scaler, and a random seed of 13 ensured reproducibility.  
We adopted a cross-silo federated learning framework with all ten clients participating ($\rho$ = 1). Models were trained in PyTorch using the Adam optimizer with a batch size of 64, learning rate of 0.005, and 100 global rounds (each comprising 5 local epochs). Each client’s dataset was split 80:20 into training and testing sets. Performance was assessed via overall accuracy, macro precision, macro recall, macro F1-score, and per-round execution time (in seconds) across all clients. Sentinel is evaluated under diverse data distributions using two recent datasets, IoTID20 \cite{ullah2020scheme} and 5GNIDD \cite{xtep-hv36-22}, which capture modern network traffic patterns. The experiments cover both IID and non-IID scenarios generated using the Dirichlet distribution \cite{hsu2019measuring}. Table~\ref{tab:params} lists the baseline hyperparameters used for comparison with \textit{Sentinel}.

\subsection{Learning Model Details}  
We employed a deep neural network (DNN) architecture inspired by prior works~\cite{meidan2018n, rey2022federated, 10492991} to evaluate Sentinel and other methods. The model consists of an input layer processing the dataset features, followed by a feature extraction module with two fully connected layers (64 and 32 neurons, each with ReLU activation). A classifier module follows, comprising a linear layer (32 to 16 neurons, ReLU activation) and an output layer mapping from 16 neurons to the number of classes.  
This architecture is utilized in Sentinel-I for both teacher and student models. 

As shown in Table \ref{tab:sentinel_architectures}, Sentinel comprises three variants. Sentinel-I employs identical architectures for teacher and student models but differentiates them through distinct training processes. Sentinel-II enhances the teacher model by expanding its feature extraction and classification layers while retaining the original student model. Sentinel-III introduces a hybrid CNN-MLP architecture, where the CNN extracts features, and the MLP performs multiclass classification. To reduce communication overhead, the student model in Sentinel-II and Sentinel-III is smaller than the teacher model. Notably, during inference, the teacher model is used for final predictions.

\subsection{Results Analysis and Discussion}
\textcolor{black}{The comprehensive evaluation, with results detailed in Tables \ref{IoT_ID_table} and \ref{5GNIDD_table}, unequivocally establishes the superiority of the \textit{Sentinel} framework. Across both the complex, multi-attack IoTID20 dataset and the large-scale 5GNIDD dataset, all \textit{Sentinel} variants consistently outperform state-of-the-art federated learning methods. The performance gap is most significant in the challenging non-IID scenarios, highlighting \textit{Sentinel}'s inherent robustness to the data heterogeneity.
\subsubsection{Dominance Under Extreme Data Heterogeneity ($\alpha = 0.1$)}
The true test of a FL system lies in its resilience to client data heterogeneity.  In this extreme non-IID scenario, the limitations of existing approaches become starkly apparent.}

\textcolor{black}{On the IoTID20 dataset, conventional methods like FedAvg and Ditto collapse, with their macro F1-scores plummeting to an unusable 13.07\%. Even methods designed to handle heterogeneity, such as Scaffold and FedProx, struggle immensely, achieving only 11.48\% and 31.36\%, respectively. This demonstrates that simple parameter averaging or basic personalization is fundamentally insufficient to overcome the feature and label skew inherent in FL.}

\textcolor{black}{In stark contrast, \textit{Sentinel} thrives under this adversity. Sentinel-III achieves a remarkable 78.22\% macro F1-score, outperforming the next-best personalized competitor, FedRep (61.29\%), by a commanding ~17 percentage points. This dominance is mirrored on the 5GNIDD dataset, where Sentinel-II attains a 97.29\% F1-score, decisively surpassing strong baselines like Scaffold (88.78\%) and APPLE (88.90\%).This underscores that \textit{Sentinel}'s architecture, which synergizes personalized distillation with explicit feature alignment, and class-balanced loss is fundamentally more robust to data heterogeneity.}
\begin{table}
\centering
\caption{Model Architectures Used in Sentinel Variants}
\label{tab:sentinel_architectures}
\resizebox{0.48\textwidth}{!}{%
\begin{tabular}{@{} c|c|c @{}}
\toprule
\textbf{Variant} & \textbf{Teacher Model} & \textbf{Student Model} \\ 
\midrule
\textbf{Sentinel-I} & \makecell{FE: Linear(64,32) \\ CL: Linear(32,16)} &
\makecell{FE: Linear(64,32) \\ CL: Linear(32,16)} \\ 
\midrule
\textbf{Sentinel-II} & \makecell{FE: Linear(128,64) \\ CL: Linear(64,32)} &
\makecell{FE: Linear(64,32) \\ CL: Linear(32,16)} \\ 
\midrule
\textbf{Sentinel-III} & \makecell{CNN: Conv1D(64), Conv1D(128) \\ MLP: Linear(256,128)} &
\makecell{CNN: Conv1D(32), Conv1D(64) \\ MLP: Linear(128,64)} \\ 
\bottomrule
\end{tabular}%
}
\begin{tablenotes}
\footnotesize
\item[] \textit{Note:}  FE = Feature Extractor; CL = Classifier
\end{tablenotes}
\end{table}
\begin{table}
\centering
\caption{\textcolor{black}{Baseline Implementation Parameters}}
\label{tab:params}
\begin{threeparttable}
\renewcommand{\arraystretch}{1.05}
\resizebox{0.48\textwidth}{!}{%
\begin{tabular}{l p{2cm} p{5.2cm}}
\toprule
\textbf{Method} & \textbf{Parameter} & \textbf{Value / Description} \\
\midrule
\textbf{FedAvg} \cite{mcmahan2017communication}
& -- & Standard Federated Averaging baseline. \\ 
\midrule
\textbf{FedProx} \cite{li2020federated}
& $\mu$ & 0.01; proximal-term coefficient (regularizes local updates toward global model). \\
\midrule
\textbf{SCAFFOLD} \cite{karimireddy2020scaffold}
& Step size ($\lambda$) & 1; controls drift correction via control variates. \\
\midrule
\textbf{PerFedAvg} \cite{fallah2020personalized}
& -- & Default personalized FL baseline. \\
\midrule
\textbf{pFedMe} \cite{t2020personalized}
& \makecell[l]{$K$\\ $\lambda$\\ $\beta$\\ $\eta_p$} 
& \makecell[l]{5; local personalized steps per batch.\\
0.15; regularization strength.\\
1.0; momentum for aggregation.\\
0.01; personalized learning rate (inner-loop).} \\
\midrule
\textbf{Ditto} \cite{li2021ditto}
& \makecell[l]{$\mu$\\ $E_p$} 
& \makecell[l]{0.5; personalization regularization weight.\\
1; local epochs for personalization.} \\
\midrule
\textbf{FedRep} \cite{collins2021exploiting}
& $E_p$ & 10; personalized head training epochs. \\
\midrule
\textbf{BalanceFL} \cite{shuai2022balancefl}
& \makecell[l]{Temperature $T$\\ $\lambda$} 
& \makecell[l]{2; temperature for class-balancing.\\
1; balancing hyperparameter.} \\
\midrule
\textbf{APPLE} \cite{luo2022adapt}
& \makecell[l]{$\mu$\\ $\eta_2$\\ $L$} 
& \makecell[l]{0.1; proximal coefficient.
\\
0.01; learning rate for relationship vector.\\
0.5; annealing duration (proportion of rounds).} \\
\midrule
\textbf{FedKD} \cite{wu2022communication}
& \makecell[l]{Teacher Model\\ Student Model} 
& \makecell[l]{64 $\rightarrow$ 32 $\rightarrow$ 16 $\rightarrow$ \#classes\\
64 $\rightarrow$ 32 $\rightarrow$ 16 $\rightarrow$ \#classes} \\
\midrule
\textbf{FedALA} \cite{zhang2023fedala}
& \makecell[l]{rand\_percent\\ layer\_idx} 
& \makecell[l]{80; random selection percentage.\\
2; selected layer index.} \\
\midrule
\makecell[l]{\textbf{Common}\\ \textbf{Parameters}} 
& \makecell[l]{Clients $N$\\ Global rounds $R$\\ Local epochs $E$\\ Optimizer\\ DL model} 
& \makecell[l]{10 (all participating).\\
100.\\
5.\\
Adam, LR = 0.005.\\
64 $\rightarrow$ 32 $\rightarrow$ 16 $\rightarrow$ \#classes.} \\
\bottomrule
\end{tabular}%
}
\end{threeparttable}
\end{table}
\begin{table*}[ht]
\centering
\caption{Performance Comparison of Fed-IDSs on IoTID20 Dataset Under Varying Data Distribution Conditions.}
\label{IoT_ID_table}
\begin{adjustbox}{max width=0.9\textwidth}
\small
\begin{tabular}{clccccc}
\toprule
\textbf{Federated Methods} & \textbf{Scenario} & \textbf{Overall Accuracy (\%)} & \textbf{Macro Precision (\%)} & \textbf{Macro Recall (\%)} & \textbf{Macro F1-score (\%)} & \textbf{Time/Round (s)} \\
\midrule
\multirow{3}{*}{%
  \makecell[c]{\textbf{\textcolor{black}{FedAvg}}\\
  \cite{mcmahan2017communication}}
}
    & $\alpha = 0.1$ & 31.01 $\pm$ 39.44 &  11.80 $\pm$ 16.66 & 23.50 $\pm$ 14.61& 13.07 $\pm$ 16.87 \\
    & $\alpha = 1$ & 74.98 $\pm$ 18.92& 63.79 $\pm$ 09.63 & 64.80 $\pm$ 03.35  & 57.38 $\pm$ 08.23 & 76\\
    & IID  & 88.02 $\pm$ 01.09 & 88.05 $\pm$ 02.39 & 78.47 $\pm$ 02.12& 78.76 $\pm$ 02.46 \\
\midrule
\multirow{3}{*}{\makecell{\textbf{\textcolor{black}{FedProx}}\\ \cite{li2020federated}}}
    & $\alpha = 0.1$ & 55.93 $\pm$ 29.62 & 39.02 $\pm$ 16.58  & 37.36 $\pm$ 15.18 & 31.36 $\pm$ 13.60 \\
    & $\alpha = 1$   & 74.95 $\pm$ 23.54 & 67.47 $\pm$ 07.84 & 65.58 $\pm$ 06.74  & 58.28 $\pm$ 10.89 & 85 \\
    & IID  & 92.55 $\pm$ 01.56 & 88.31 $\pm$ 02.46 & 86.79 $\pm$ 03.09 & 86.97 $\pm$ 03.23  \\
\midrule

\multirow{3}{*}{\makecell{\textbf{\textcolor{black}{Scaffold}} \\ \cite{karimireddy2020scaffold}}}
    & $\alpha = 0.1$ & 31.09 $\pm$ 39.38 & 18.01 $\pm$ 19.07  & 21.84 $\pm$ 12.21 & 11.48 $\pm$ 13.62 \\
    & $\alpha = 1$   & 92.62 $\pm$ 01.90& 82.60 $\pm$ 06.95 & 76.51 $\pm$ 07.11  &76.39 $\pm$ 08.34 & 75 \\
    & IID  & 93.27 $\pm$ 00.78& 89.73 $\pm$ 01.31 & 88.00$\pm$ 02.82 & 88.21 $\pm$ 02.55  \\
\midrule
\multirow{3}{*}{\makecell{\textbf{\textcolor{black}{Per-FedAvg}} \\ \cite{fallah2020personalized}}}
    & $\alpha = 0.1$ & 94.18 $\pm$ 09.44 & 65.09 $\pm$ 13.24 & 56.97$\pm$ 14.82 & 58.51$\pm$ 15.14  \\
    & $\alpha = 1$   & 93.75 $\pm$ 02.16 &  87.42 $\pm$ 04.92  & 79.59 $\pm$ 06.25 & 80.75 $\pm$ 06.02 & 163   \\
    & IID  & 93.28 $\pm$ 00.57 & 89.41 $\pm$ 00.77  & 88.60 $\pm$ 02.31 & 88.52 $\pm$ 01.92 \\
\midrule
\multirow{3}{*}{\makecell{\textbf{\textcolor{black}{pFedME}} \\ \cite{t2020personalized}}}
    & $\alpha = 0.1$ & 35.39 $\pm$ 39.69 & 34.93 $\pm$ 19.27 & 28.83 $\pm$ 14.19 & 21.09 $\pm$ 15.14  \\
    & $\alpha = 1$   & 73.92 $\pm$ 25.75 &  70.39 $\pm$ 11.17  & 64.11 $\pm$ 09.21 & 59.07 $\pm$ 12.64 & 376  \\
    & IID  & 91.44 $\pm$ 00.21 & 87.42 $\pm$ 01..47  & 81.41 $\pm$ 00.96 & 81.46 $\pm$ 01.00 \\
\midrule
\multirow{3}{*}{\makecell{\textbf{\textcolor{black}{Ditto}} \\ \cite{li2021ditto}}}
    & $\alpha = 0.1$ & 31.01 $\pm$ 39.44 & 11.80 $\pm$ 16.66 & 23.50 $\pm$ 14.61 & 13.07 $\pm$ 16.87  \\
    & $\alpha = 1$   & 73.62 $\pm$ 24.69 &  65.19 $\pm$ 11.99  & 68.69 $\pm$ 08.63 & 60.02 $\pm$ 13.93 &  123  \\
    & IID  & 92.02 $\pm$ 00.94 & 87.63 $\pm$ 00.24  & 83.51 $\pm$ 02.06 & 83.64 $\pm$ 01.80 \\
\midrule
\multirow{3}{*}{\makecell{\textbf{\textcolor{black}{FedRep}} \\ \cite{collins2021exploiting}}}
    & $\alpha = 0.1$ & 93.01 $\pm$ 10.68 & 73.58 $\pm$ 15.08 & 59.61$\pm$ 08.12 & 61.29 $\pm$ 07.78  \\
    & $\alpha = 1$   & 91.72 $\pm$ 02.71 &  79.98 $\pm$ 07.57  & 72.61 $\pm$ 07.55 & 74.56 $\pm$ 07.20 & 75 \\
    & IID  & 91.69 $\pm$ 00.19 & 88.74 $\pm$ 00.65 & 81.17 $\pm$ 00.61 & 81.45 $\pm$ 00.58 \\
\midrule
\multirow{3}{*}{\makecell{\textbf{\textcolor{black}{BalanceFL}} \\ \cite{shuai2022balancefl}}}
    & $\alpha = 0.1$ & 28.55 $\pm$ 21.55 & 22.19 $\pm$ 13.86&30.66 $\pm$ 16.73&16.28 $\pm$ 12.47 \\
    & $\alpha = 1$   & 69.87 $\pm$ 24.97 & 65.38 $\pm$ 11.58 & 60.00 $\pm$ 08.19 &52.65$\pm$ 10.51 & \textbf{461}\\
    & IID & 91.36 $\pm$ 02.49 & 89.26 $\pm$ 02.25 & 84.45 $\pm$ 03.86& 85.95 $\pm$ 04.01 \\
\midrule
\multirow{3}{*}{\makecell{\textbf{\textcolor{black}{APPLE}} \\ \cite{luo2022adapt}}}
    & $\alpha = 0.1$ & 91.19 $\pm$ 17.54 & 61.08 $\pm$ 17.89 & 53.00 $\pm$ 17.41 & 54.75 $\pm$ 18.83  \\
    & $\alpha = 1$   & 94.05 $\pm$ 01.71 &  85.09 $\pm$ 05.56  & 80.45 $\pm$ 06.86 & 81.41 $\pm$ 06.44 & 409 \\
    & IID  & 94.17 $\pm$ 00.85 & 91.18 $\pm$ 02.34 & 89.41 $\pm$ 01.95 & 89.97 $\pm$ 01.93 \\
\midrule
\multirow{3}{*}{\makecell{\textbf{\textcolor{black}{FedKD}} \\ \cite{wu2022communication}}}
    & $\alpha = 0.1$ & 93.75 $\pm$ 07.66 & 58.97 $\pm$ 16.16 & 51.32 $\pm$ 17.01 & 52.11 $\pm$ 17.05 \\
    & $\alpha = 1$   & 90.56 $\pm$ 02.07 &  77.69 $\pm$ 13.01  & 62.83 $\pm$ 11.05 & 65.10 $\pm$ 11.41 & 311 \\
    & IID  & 90.00 $\pm$ 00.49 & 90.18 $\pm$ 00.82  & 76.14 $\pm$ 01.36 & 76.94 $\pm$ 01.18 \\
\midrule
\multirow{3}{*}{\makecell{\textbf{\textcolor{black}{FedKD}} \\(\textcolor{black}{Without Decomposition}) }}
    & $\alpha = 0.1$ & 94.08 $\pm$ 07.65 & 64.78 $\pm$ 14.47 & 57.11 $\pm$ 11.16 & 58.59 $\pm$ 10.68 \\
    & $\alpha = 1$   & 91.90 $\pm$ 02.08 &  83.37 $\pm$ 08.04  & 69.78 $\pm$ 09.26 & 72.45 $\pm$ 08.88 & 306 \\
    & IID  & 91.26 $\pm$ 00.20 & 89.00 $\pm$ 01.27 & 80.39 $\pm$ 00.89 & 81.64 $\pm$ 00.96 \\
\midrule
\multirow{3}{*}{\makecell{\textbf{\textcolor{black}{FedALA}} \\ \cite{zhang2023fedala}}}
    & $\alpha = 0.1$ & 74.90 $\pm$ 27.49 & 42.30 $\pm$ 11.00 & 49.39 $\pm$ 11.29 & 39.39 $\pm$ 13.07  \\
    & $\alpha = 1$   & 84.40 $\pm$ 12.96 &  74.18$\pm$ 12.17 & 70.23 $\pm$ 06.82 & 67.78$\pm$ 08.18 & 92  \\
    & IID  & 92.33 $\pm$ 00.49 & 87.07 $\pm$ 01.25  & 86.15 $\pm$ 00.68 & 85.97 $\pm$ 00.89 \\
\midrule
\multirow{3}{*}
{\makecell{\textbf{\textcolor{black}{Sentinel-I}} \\(\textcolor{black}{Same $M_T$ and $M_S$ DNN models})}}
& $\alpha = 0.1$ & 95.62 $\pm$ 10.03 & 83.39 $\pm$ 11.48 & 76.75 $\pm$ 13.26 & 77.92 $\pm$ 12.42 \\
& $\alpha = 1$   & 96.28 $\pm$ 00.82 &  89.64 $\pm$ 05.52  & 87.08 $\pm$ 07.45 & 87.61 $\pm$ 07.03 & 297 \\
& IID  & 95.27 $\pm$ 00.86 & 92.96 $\pm$ 01.01 & 92.08 $\pm$ 01.43 & 92.42 $\pm$ 01.28\\\midrule
\multirow{3}{*}
{\makecell{\textbf{\textcolor{black}{Sentinel-II}} \\(\textcolor{black}{Strong $M_T$ and light $M_S$ DNN models})}}
& $\alpha = 0.1$ & \textbf{96.00 $\pm$ 09.24} &83.36 $\pm$ 11.57 & 76.74 $\pm$ 13.43 & 78.13 $\pm$ 12.71 \\
& $\alpha = 1$   & \textbf{96.64 $\pm$ 00.70} &  \textbf{91.41 $\pm$ 04.81}  & \textbf{88.40 $\pm$ 06.75} & \textbf{88.78 $\pm$ 06.16} & 325 \\
& IID  & \textbf{95.99 $\pm$ 00.24} & \textbf{93.80 $\pm$ 00.61} & \textbf{93.47 $\pm$ 00.38} & \textbf{93.59 $\pm$ 00.25}\\\midrule
\multirow{3}{*}
{\makecell{\textbf{\textcolor{black}{Sentinel-III}} \\(\textcolor{black}{Strong $M_T$ and light $M_S$ CNN models})}}
& $\alpha = 0.1$ & 95.65 $\pm$ 10.32 & \textbf{83.78 $\pm$ 11.78} & \textbf{77.21 $\pm$ 13.48} & \textbf{78.22 $\pm$ 12.90} \\
& $\alpha = 1$   & 95.72 $\pm$ 00.97 &  88.39 $\pm$ 05.15  & 86.25 $\pm$ 07.51 & 85.75 $\pm$ 07.01 & 550 \\
& IID  & 95.47 $\pm$ 00.45 & 93.42 $\pm$ 00.74 & 92.46 $\pm$ 00.39 & 92.83 $\pm$ 00.51\\
\bottomrule
\end{tabular}
\end{adjustbox}
\footnotesize
\begin{minipage}{0.9\textwidth}
\vspace{3pt} 
\raggedright\hangindent=1em 
\textit{Note: Each metric value represents the mean and standard deviation across results from 10 clients.}
\end{minipage}
\end{table*}

\begin{table*}[ht]
\centering
\caption{Performance Comparison of Fed-IDSs on 5GNIDD Dataset Under Varying Data Distribution Conditions.}
\label{5GNIDD_table}
\begin{adjustbox}{max width=0.95\textwidth}
\small
\begin{tabular}{clccccc}
\toprule
\textbf{Federated Methods} & \textbf{Scenario} & \textbf{Overall Accuracy (\%)} & \textbf{Macro Precision (\%)} & \textbf{Macro Recall (\%)} & \textbf{Macro F1-score (\%)} & \textbf{Time/Round (s)} \\
\midrule
\multirow{3}{*}{%
  \makecell[c]{\textbf{\textcolor{black}{FedAvg}}\\
               \cite{mcmahan2017communication}}
}
    & $\alpha = 0.1$ &41.71 $\pm$ 34.08& 21.50$\pm$ 16.68&24.00 $\pm$ 16.02& 16.57 $\pm$ 18.31 \\
    & $\alpha = 1$ & 93.72 $\pm$ 05.49 & 84.90 $\pm$ 02.79&83.95 $\pm$ 02.76&83.45 $\pm$ 03.19 & 152 \\
    & IID & 99.53 $\pm$ 00.03&86.95 $\pm$ 00.15 &87.02 $\pm$ 00.23& 86.85 $\pm$ 00.18\\
\midrule
\multirow{3}{*}{\makecell{\textbf{\textcolor{black}{FedProx}}\\ \cite{li2020federated}}}
    & $\alpha = 0.1$ & 75.44 $\pm$ 28.04 & 44.49 $\pm$ 16.78  & 53.16 $\pm$ 20.54 & 41.56 $\pm$ 18.06 \\
    & $\alpha = 1$   & 95.13 $\pm$ 03.21 & 93.03 $\pm$ 06.33 & 95.74 $\pm$ 05.58  & 93.85 $\pm$ 06.40 & 161  \\
    & IID  & 99.82 $\pm$ 00.02 & 99.70 $\pm$ 00.08 & 99.68 $\pm$ 00.08 & 99.69 $\pm$ 00.08  \\
\midrule
\multirow{3}{*}{\makecell{\textbf{\textcolor{black}{Scaffold}} \\ \cite{karimireddy2020scaffold}}}
    & $\alpha = 0.1$ & 99.85 $\pm$ 00.14 & 89.01 $\pm$ 12.63 & 89.20 $\pm$ 13.85& 88.78 $\pm$ 13.56 \\
    & $\alpha = 1$   & 99.76 $\pm$ 00.09& 99.39 $\pm$ 00.44&  99.64 $\pm$ 00.06& \textbf{99.51 $\pm$ 00.23} & 146\\
    & IID &  99.82 $\pm$ 00.02 & 99.70 $\pm$ 00.07& 99.70 $\pm$ 00.07 & 99.70 $\pm$ 00.07\\
\midrule
\multirow{3}{*}{\makecell{\textbf{\textcolor{black}{Per-FedAvg}} \\ \cite{fallah2020personalized}}}
    & $\alpha = 0.1$ & 99.78 $\pm$ 00.26 & 76.81 $\pm$ 14.29 & 78.23$\pm$ 14.15 & 76.69$\pm$ 14.04  \\
    & $\alpha = 1$   & 99.82 $\pm$ 00.07 &  99.47 $\pm$ 00.28  & 99.53 $\pm$ 00.29 & 99.50 $\pm$ 00.23  & 310 \\
    & IID  & 99.82 $\pm$ 00.02 & 99.67 $\pm$ 00.13  & 99.69 $\pm$ 00.07 & 99.68 $\pm$ 00.09 \\
\midrule
\multirow{3}{*}{\makecell{\textbf{\textcolor{black}{pFedME}} \\ \cite{t2020personalized}}}
    & $\alpha = 0.1$ & 40.17 $\pm$ 35.99 & 20.84 $\pm$ 14.96 & 28.69 $\pm$ 23.00  & 16.76 $\pm$ 17.61  \\
    & $\alpha = 1$   & 92.91 $\pm$ 05.15 &  93.25 $\pm$ 02.41  & 96.08 $\pm$ 01.12 & 93.68 $\pm$ 02.30 & 709  \\
    & IID  & 99.59 $\pm$ 00.03 & 98.82 $\pm$ 00.08  & 98.06 $\pm$ 00.22 & 98.37 $\pm$ 00.16 \\
\midrule
\multirow{3}{*}{\makecell{\textbf{\textcolor{black}{Ditto}} \\ \cite{li2021ditto}}}
    & $\alpha = 0.1$ &44.99 $\pm$ 38.15 & 35.46 $\pm$ 22.30 & 32.57 $\pm$ 19.96 & 24.94 $\pm$ 23.69  \\
    & $\alpha = 1$   & 93.78 $\pm$ 05.12 &  83.61 $\pm$ 03.95  & 84.55 $\pm$ 03.80 & 82.66 $\pm$ 05.44  & 233  \\
    & IID  & 99.69 $\pm$ 00.03 & 99.76 $\pm$ 00.12  & 98.03 $\pm$ 00.25 & 98.84 $\pm$ 00.15 \\
\midrule
\multirow{3}{*}{\makecell{\textbf{\textcolor{black}{FedRep}} \\ \cite{collins2021exploiting}}}
    & $\alpha = 0.1$ & 99.10 $\pm$ 01.16 & 72.56 $\pm$ 16.13 & 70.81 $\pm$ 15.39 & 71.07 $\pm$ 15.32  \\
    & $\alpha = 1$   & 99.68 $\pm$ 00.09 &  99.07 $\pm$ 00.35  & 97.05 $\pm$ 01.84 & 97.83 $\pm$ 01.35 & 142 \\
    & IID  & 99.77 $\pm$ 00.02 & 99.51 $\pm$ 00.19& 99.33 $\pm$ 00.27 & 99.41 $\pm$ 00.23 \\
\midrule
\multirow{3}{*}{\makecell{\textbf{\textcolor{black}{BalanceFL}} \\ \cite{shuai2022balancefl}}}
    & $\alpha = 0.1$ & 27.01 $\pm$ 16.44& 21.44 $\pm$ 07.30& 25.90 $\pm$ 10.78& 12.40 $\pm$ 06.72\\
    & $\alpha = 1$   & 85.91 $\pm$ 07.68&80.46 $\pm$ 09.64& 82.22 $\pm$ 06.07&77.11 $\pm$ 08.99 & \textbf{1385} \\
    & IID & 99.26 $\pm$ 00.11 & 98.84 $\pm$ 00.16& 99.47 $\pm$ 00.07 & 99.15 $\pm$ 00.09  \\
\midrule
\multirow{3}{*}
{\makecell{\textbf{\textcolor{black}{APPLE}} \\ \cite{luo2022adapt}}}
    & $\alpha = 0.1$ & 99.88 $\pm$ 00.10 & 88.72 $\pm$ 12.20 & 90.48 $\pm$ 11.46 & 88.90 $\pm$ 11.99  \\
    & $\alpha = 1$   & 99.82 $\pm$ 00.07 &  99.62 $\pm$ 00.14  & 99.37 $\pm$ 00.58 & 99.49 $\pm$ 00.36  & 832 \\
    & IID  & 99.82 $\pm$ 00.02 & 99.67 $\pm$ 00.14  & 99.72 $\pm$ 00.05 & 99.69 $\pm$ 00.09\\
\midrule
\multirow{3}{*}{\makecell{\textbf{\textcolor{black}{FedKD}} \\ \cite{wu2022communication}}}
    & $\alpha = 0.1$ & 95.26 $\pm$ 05.66 & 51.54 $\pm$ 23.77 & 43.18 $\pm$ 25.50 & 43.16 $\pm$ 25.73 \\
    & $\alpha = 1$   & 96.43 $\pm$ 01.25 & 60.50 $\pm$ 10.67 & 59.11 $\pm$ 08.89 &  58.88 $\pm$ 08.80& 605\\
    & IID  &  95.76 $\pm$ 00.59 & 74.27 $\pm$ 08.88 & 60.69 $\pm$ 06.59 & 61.76 $\pm$ 07.57 \\
\midrule
\multirow{3}{*}{\makecell{\textbf{\textcolor{black}{FedKD}} \\(\textcolor{black}{Without Decomposition})}}
    & $\alpha = 0.1$ & 95.43 $\pm$ 05.07 & 54.22 $\pm$ 23.00 & 47.74 $\pm$ 24.71 & 47.68 $\pm$ 24.80 \\
    & $\alpha = 1$   & 96.46 $\pm$ 02.08 &  78.75 $\pm$ 04.71  & 75.26 $\pm$ 05.72 & 76.45 $\pm$ 05.43 & 592 \\
    & IID  & 99.47 $\pm$ 00.06 & 87.95 $\pm$ 00.35 & 87.21 $\pm$ 00.71 & 87.54 $\pm$ 00.56 \\
\midrule
\multirow{3}{*}{\makecell{\textbf{\textcolor{black}{FedALA}} \\ \cite{zhang2023fedala}}}
    & $\alpha = 0.1$ & 87.00 $\pm$ 11.75 & 45.08 $\pm$ 14.48 & 50.97 $\pm$ 13.27 & 45.13 $\pm$ 14.56  \\
    & $\alpha = 1$   & 97.58 $\pm$ 03.01 &  87.47 $\pm$ 00.91 & 85.07 $\pm$ 04.89 & 85.69 $\pm$ 04.22 & 166 \\
    & IID  & 99.78 $\pm$ 00.02 & \textbf{99.81 $\pm$ 00.03}  & 99.56 $\pm$ 00.08 & 99.69 $\pm$ 00.05 \\
\midrule
\multirow{3}{*}
{\makecell{\textbf{\textcolor{black}{Sentinel-I}} \\(\textcolor{black}{Same $M_T$ and $M_S$ DNN models})}}
& $\alpha = 0.1$ & 99.95 $\pm$ 00.04 & 97.38 $\pm$ 06.53 & 96.68 $\pm$ 05.68 & 96.95 $\pm$ 06.09 & \\
& $\alpha = 1$   & 99.84 $\pm$ 00.06 &  98.18 $\pm$ 03.39  & 98.48 $\pm$ 03.31 & 98.28 $\pm$ 03.31 & 585  \\
& IID  & 99.83$\pm$ 00.03 & 99.76 $\pm$ 00.06 & 99.65 $\pm$ 00.07 & 99.71 $\pm$ 00.06 & \\\midrule

\multirow{3}{*}
{\makecell{\textbf{\textcolor{black}{Sentinel-II}} \\(\textcolor{black}{Strong $M_T$ and light $M_S$ DNN models})}}
& $\alpha = 0.1$ & \textbf{99.96 $\pm$ 00.04} & \textbf{98.12 $\pm$ 04.43} & \textbf{96.90 $\pm$ 05.71} & \textbf{97.29 $\pm$ 05.08} & \\
& $\alpha = 1$   & \textbf{99.85 $\pm$ 00.07} &  \textbf{99.56 $\pm$ 00.47}  & \textbf{99.66 $\pm$ 00.23} & \textbf{99.60 $\pm$ 00.27} & 631  \\
& IID  & \textbf{99.85 $\pm$ 00.01} & 99.73 $\pm$ 00.05 & 99.72 $\pm$ 00.09 & 99.72 $\pm$ 00.06 & \\\midrule

\multirow{3}{*}
{\makecell{\textbf{\textcolor{black}{Sentinel-III}} \\(\textcolor{black}{Strong $M_T$ and $M_S$ CNN models})}}
    & $\alpha = 0.1$ & 99.95 $\pm$ 00.04 & 96.13 $\pm$ 06.64 & 96.89 $\pm$ 05.72 & 96.20 $\pm$ 06.08 & \\
    & $\alpha = 1$   & 99.81 $\pm$ 00.09 & \textbf{99.63 $\pm$ 00.15} & 99.54 $\pm$ 00.37 & 99.58 $\pm$ 00.24 & 950\\
    & IID  & 99.83 $\pm$ 00.04 & 99.72 $\pm$ 00.09 & \textbf{99.74 $\pm$ 00.06} & \textbf{99.73 $\pm$ 00.07} & \\
\bottomrule
\end{tabular}

\end{adjustbox}
\footnotesize
\begin{minipage}{0.9\textwidth}
\vspace{3pt} 
\raggedright\hangindent=1em 
\textit{Note: Each metric value represents the mean and standard deviation across results from 10 clients.}
\end{minipage}
\end{table*}
\begin{figure}
    \centering
    \includegraphics[width=0.4\textwidth]{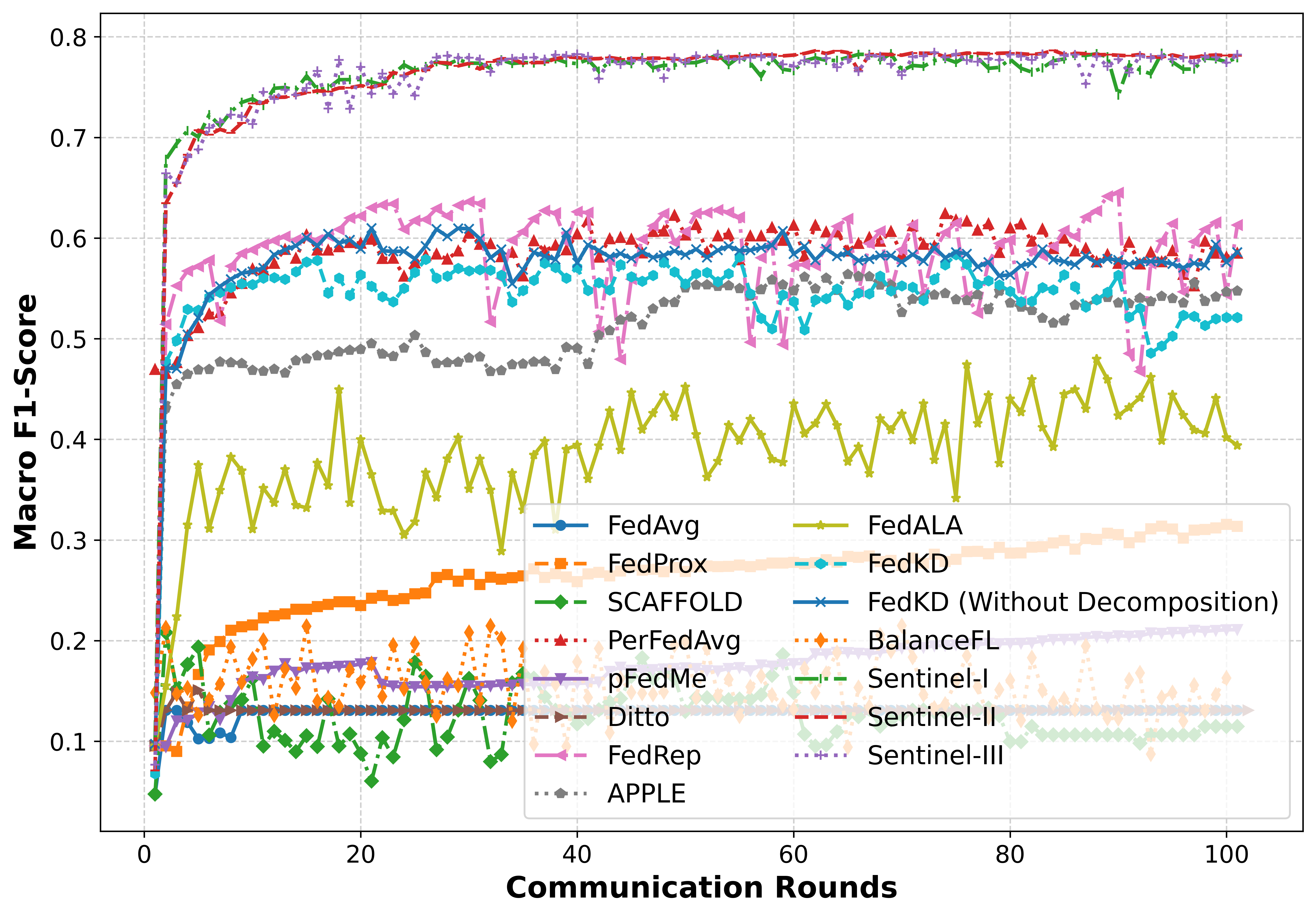}
    \caption*{\footnotesize{$(a)$}}
    \label{fig:dirichlet_01}
\vspace{1em}
    \includegraphics[width=0.4\textwidth]{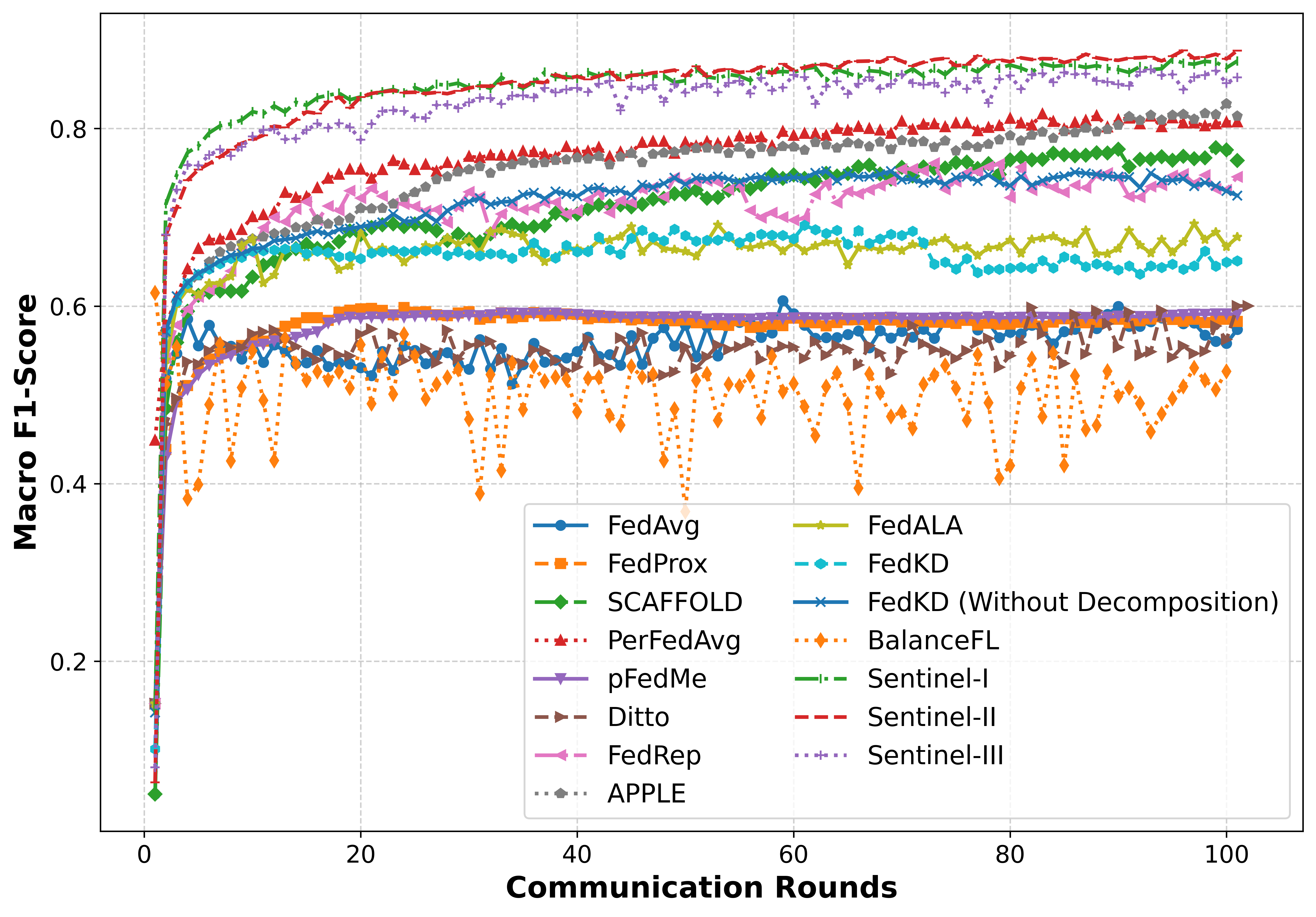}
    \caption*{\footnotesize{$(b)$}}
    \label{fig:dirichlet_1}
\vspace{1em}
    \includegraphics[width=0.4\textwidth]{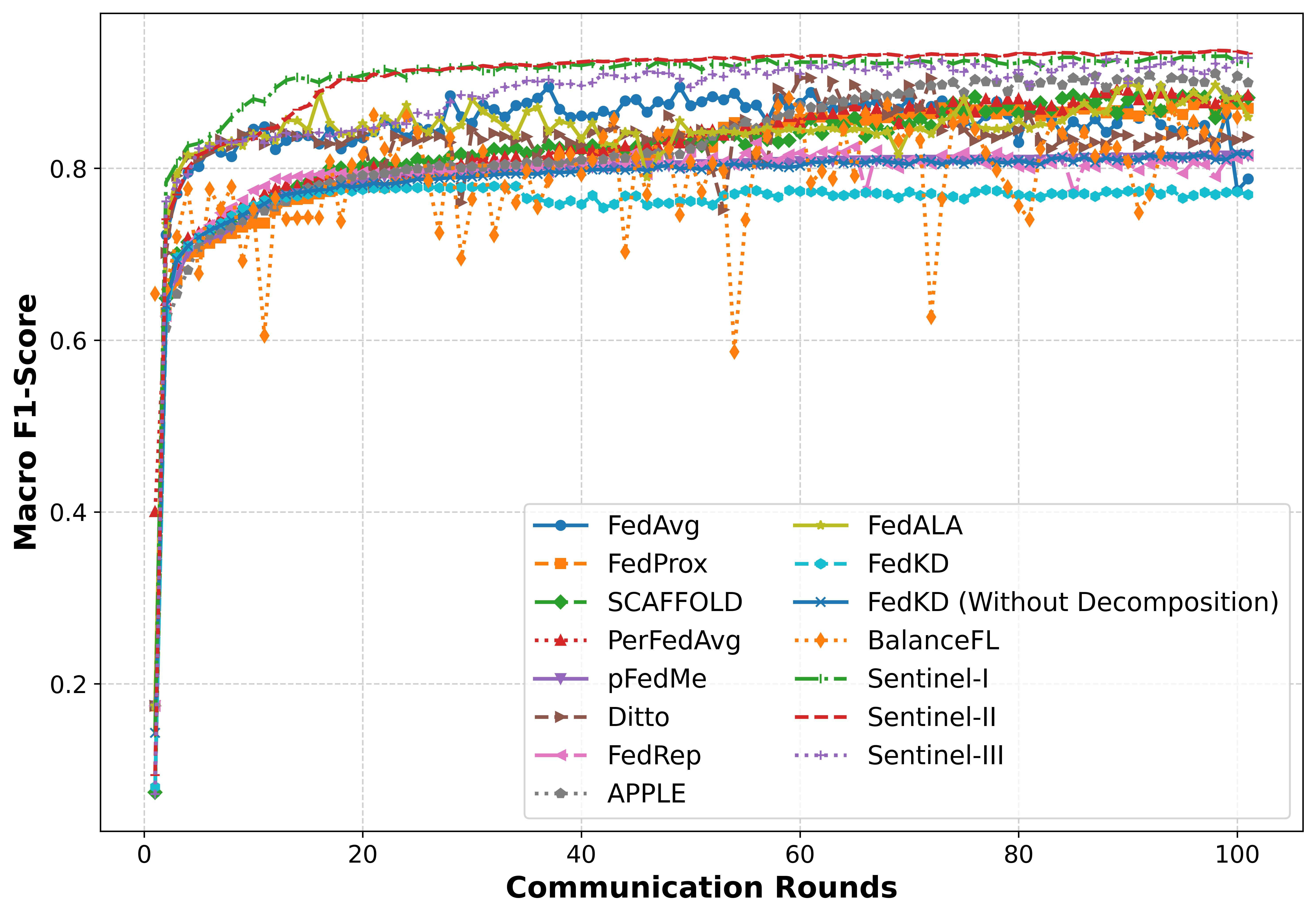}
    \caption*{\footnotesize{$(c)$}}
    \label{fig:dirichlet_iid}

    \caption{\textcolor{black}{Macro F1-score Comparison of Fed-IDSs on IoTID20 Dataset Under Varying Data Distribution Conditions.}}
    \label{fig:IoT_results}
\end{figure}
\begin{figure}
    \centering
    \includegraphics[width=0.4\textwidth]{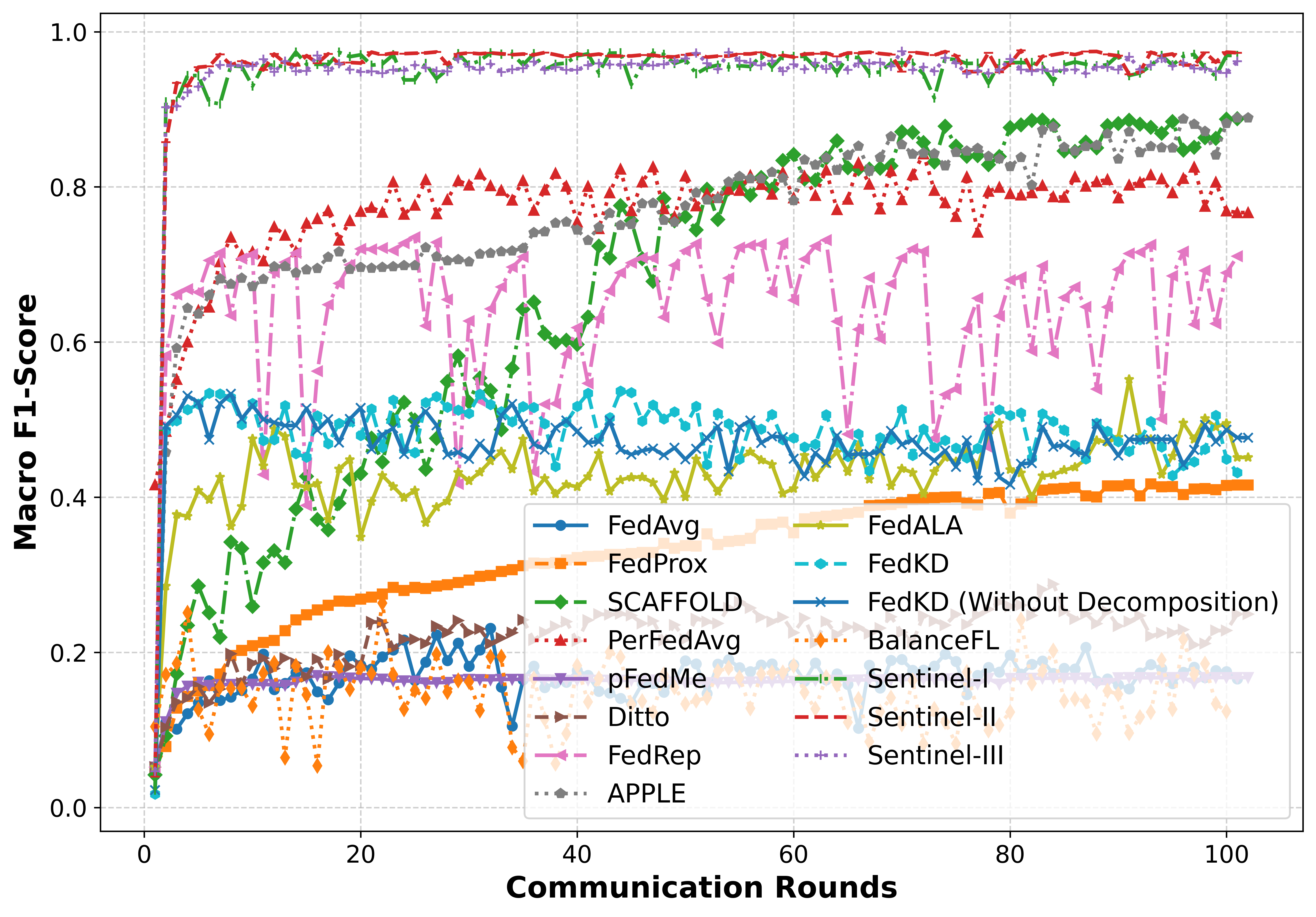}
    \caption*{\footnotesize{$(a)$}}
    \label{fig:5gnidd_01}
\vspace{1em}
    \includegraphics[width=0.4\textwidth]{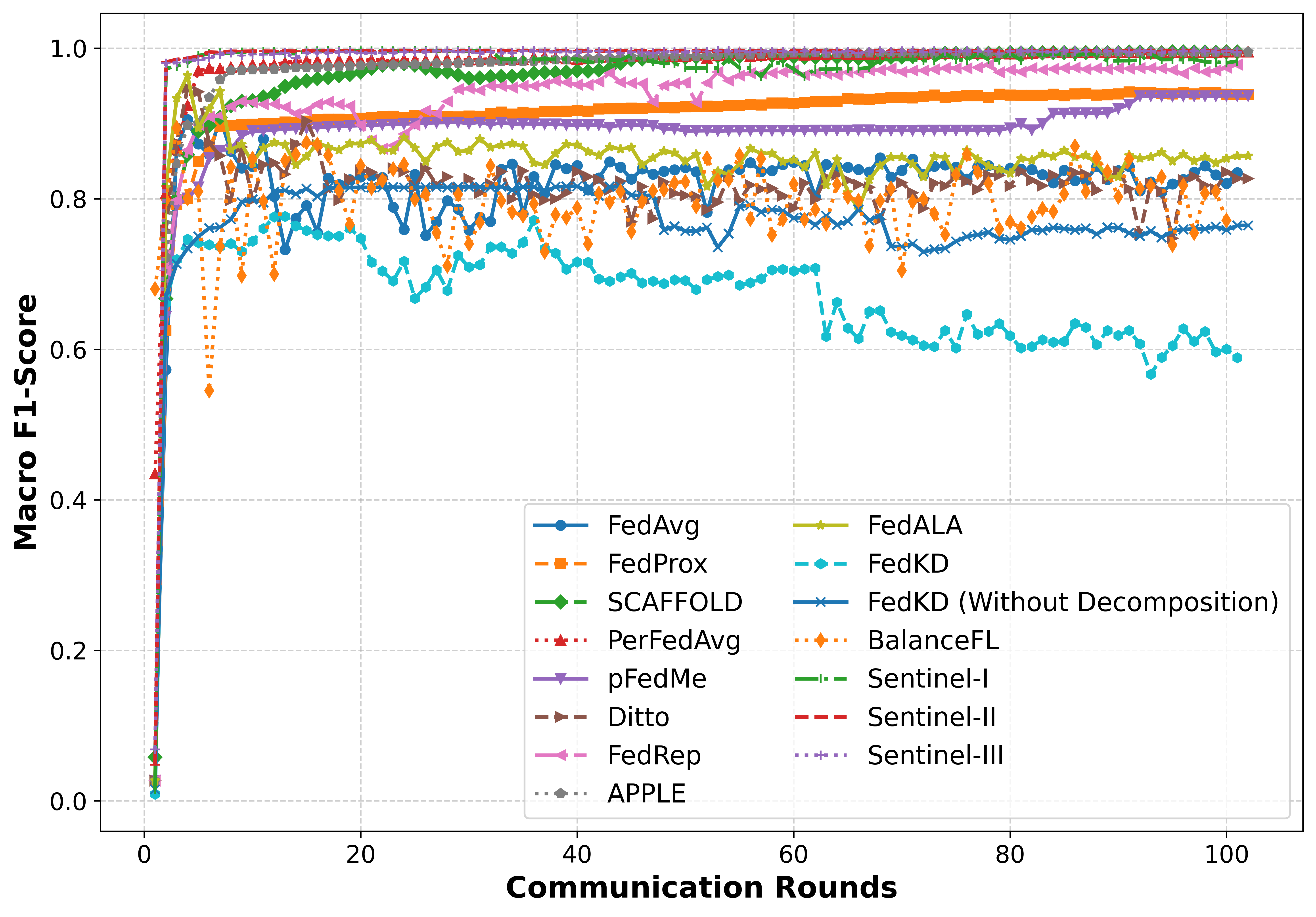}
    \caption*{\footnotesize{$(b)$}}
    \label{fig:5gnidd_1}
\vspace{1em}
    \includegraphics[width=0.4\textwidth]{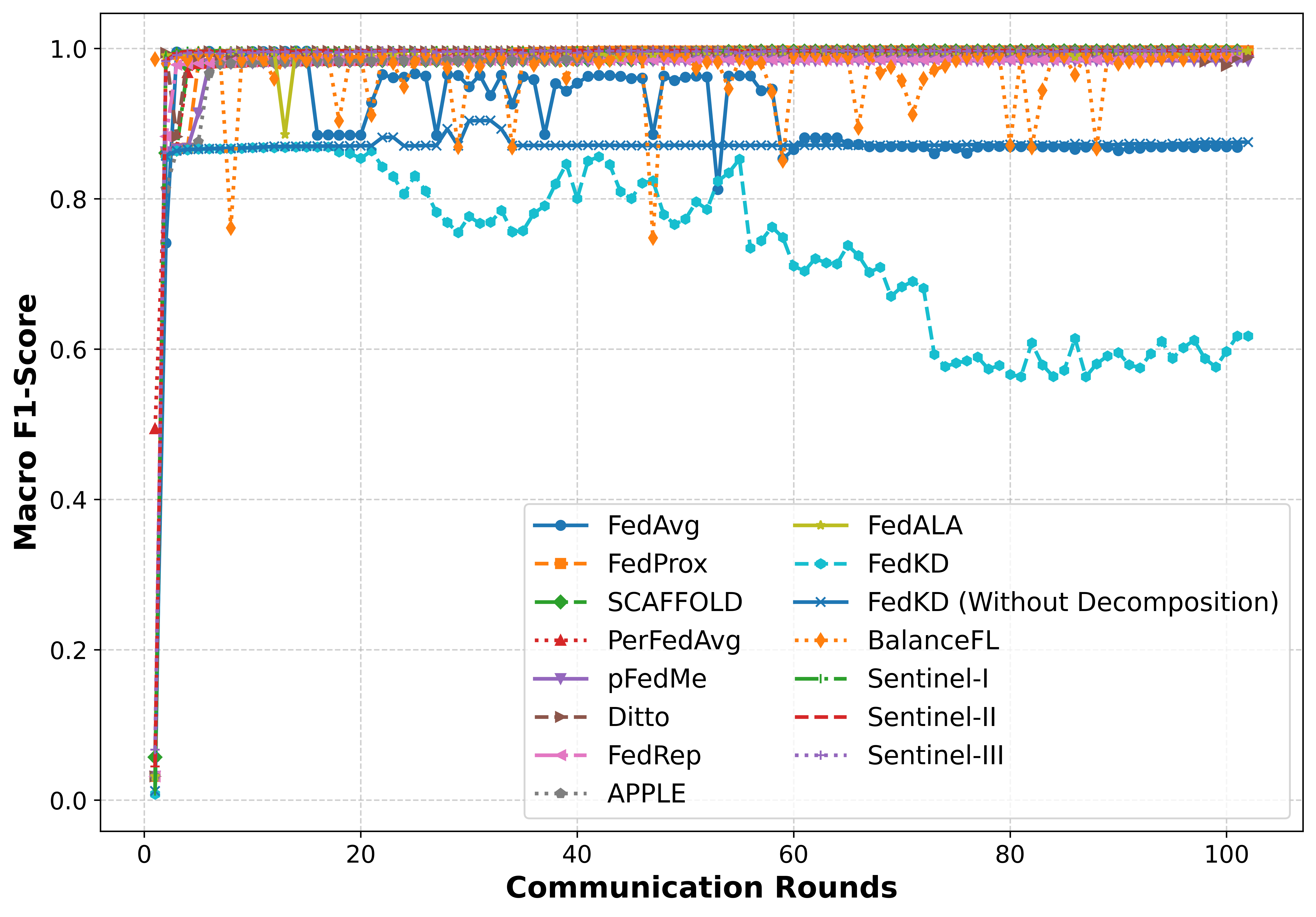}
    \caption*{\footnotesize{$(c)$}}
    \label{fig:5gnidd_iid}

    \caption{\textcolor{black}{Macro F1-score Comparison of Fed-IDSs on 5GNIDD Dataset Under Varying Data Distribution Conditions.}}
    \label{fig:5gnidd_results}
\end{figure}

\subsubsection{\textcolor{black}{Strategic Trade-offs Among Sentinel Variants}}

\textcolor{black}{We experimented with three variants of \textit{Sentinel}, as described in Table~\ref{tab:sentinel_architectures}, each consistently outperforming state-of-the-art methods while offering different balances among performance, computational efficiency, and architectural complexity.}

\paragraph{\textcolor{black}{Sentinel-I (Symmetric DNNs).}}
\textcolor{black}{This baseline variant employs identical DNNs for both the teacher and student models, enabling fair comparison with existing approaches that use the same architecture. Despite its simplicity, \textit{Sentinel-I} surpasses most competitors, demonstrating that the core distillation and alignment mechanisms are highly effective even without architectural asymmetry. Its training cost is highly competitive ($\approx 297$ s/round on IoTID20), making it more efficient than strong baselines like BalanceFL, APPLE, and FedKD. Consequently, \textit{Sentinel-I} is an ideal candidate for resource-constrained edge environments where stability and low latency are paramount.}

\paragraph{\textcolor{black}{Sentinel-II (Asymmetric DNNs).}}
\textcolor{black}{As the flagship realization of our framework, \textit{Sentinel-II} leverages a larger, more expressive teacher model for localized guidance while maintaining a lightweight student for communication-efficient aggregation. This asymmetric design captures the best of both worlds: rich, feature-level supervision during local training and minimal communication overhead during global updates. Empirically, \textit{Sentinel-II} consistently leads across all metrics, especially in non-IID settings (e.g., achieving an 88.78\,\% macro F1-score on IoTID20 at $\alpha=1$, decisively beating FedRep’s 74.56\,\%). This performance gain is achieved at a moderate computational cost ($\approx 325$ s/round), establishing \textit{Sentinel-II} as the optimal balance of accuracy, robustness, and practicality for real-world federated IDS deployments.}

\paragraph{\textcolor{black}{Sentinel-III (CNN-based Hybrid).}}
\textcolor{black}{To explore architectural impact, we also implemented a hybrid variant pairing a strong CNN-MLP teacher with a lightweight CNN-MLP student. While not directly comparable to the DNN-only baselines, \textit{Sentinel-III} confirms that our framework is versatile and can be enhanced with more expressive feature extractors. However, this comes at a significantly higher computational cost ($\approx 550$ s/round) due to deeper model architectures.}
\textcolor{black}{
\subsubsection{Superior Minority Class Detection and Computational Cost}  
A key requirement for any IDS is the reliable detection of rare yet critical attacks. The Macro Recall metric, which emphasizes performance on minority classes, underscores \textit{Sentinel}'s advantage. On the IID IoTID20 dataset, \textit{Sentinel-II} attains a Macro Recall of 93.47\%, notably surpassing specialized methods such as APPLE (89.41\%) and BalanceFL (84.45\%). This demonstrates that its class-balanced loss, coupled with robust representation learning, effectively addresses the class imbalance challenge that hampers other approaches.  }

\textcolor{black}{In terms of efficiency, while \textit{Sentinel} incurs a higher per-round training cost compared to lightweight baselines such as FedAvg, this overhead reflects its more sophisticated local optimization. Importantly, the additional cost is offset by rapid convergence, requiring fewer global rounds to reach peak performance. Combined with deployment on resource-capable MEC devices, this makes \textit{Sentinel}'s computational footprint both practical and well-suited for real-world scenarios. }

\subsubsection{\textcolor{black}{Rapid Convergence and Stability}}
\textcolor{black}{Beyond peak performance, \textit{Sentinel} also ensures stability and efficiency. As shown in Figures~\ref{fig:IoT_results} and~\ref{fig:5gnidd_results}, the \textit{Sentinel} variants not only achieve higher F1-scores but also converge significantly faster. They reach near-optimal performance within a few dozen global rounds and sustain it with minimal fluctuation. This stands in sharp contrast to the erratic and slow progress of baselines such as FedAvg and FedProx, particularly under non-IID conditions.}
\textcolor{black}{This stability is further evidenced by the low standard deviations reported in Tables~\ref{IoT_ID_table} and~\ref{5GNIDD_table}. For example, on the challenging 5GNIDD dataset ($\alpha=0.1$), \textit{Sentinel-II} exhibits an F1-score variance of just $\pm 5.08\,\%$, markedly lower than the variances observed for Ditto ($\pm 23.69\,\%$) and FedKD ($\pm 25.73\,\%$).}

These comprehensive results across multiple metrics, datasets, and data distribution scenarios validate the efficacy of Sentinel in addressing the challenges posed by non-IID data distributions in federated intrusion detection systems, and underscore its potential for practical deployment in real-world federated learning environments with heterogeneous client data. Future work could focus on further optimizing the knowledge distillation process and investigating adaptive mechanisms that automatically adjust model parameters in response to varying degrees of data heterogeneity.
\begin{table}
\centering
\caption{\footnotesize \textcolor{black}{Ablation Study of Sentinel's core components on the \textbf{IoTID20} dataset. The table shows the macro F1-Score (\%) under varying data heterogeneity ($\alpha=0.1, 1.0$) and IID conditions. The best results (\textbf{bold}) are achieved only when all three mechanisms---Class-Balanced Loss (\texttt{Bal. Task}), Bidirectional KD (\texttt{Bi-KD}), and Feature Alignment (\texttt{Align})---are active, confirming their synergistic importance}}
\label{tab:ablation_iotid20_all}
\begin{adjustbox}{max width=0.48\textwidth}
\small
\begin{tabular}{c c c c c c c}
\toprule
Task & Bal.~Task & Bi-KD & Align & F1 ($\alpha{=}0.1$) & F1 ($\alpha{=}1$) & F1 (IID) \\
\midrule
\ding{51} & \ding{55} & \ding{51} & \ding{55} 
& 69.72 $\pm$ 11.84 & 85.46 $\pm$ 7.91 & 92.22 $\pm$ 0.76 \\

\ding{55} & \ding{51} & \ding{51} & \ding{55} 
& 71.98 $\pm$ 13.40 & 86.01 $\pm$ 8.53 & 92.01 $\pm$ 1.04 \\

\ding{55} & \ding{51} & \ding{51} & \ding{51} 
& \textbf{77.92 $\pm$ 12.42} & \textbf{87.61 $\pm$ 7.03} & \textbf{92.42 $\pm$ 1.28} \\

\ding{51} & \ding{55} & \ding{51} & \ding{51} 
& 71.19 $\pm$ 10.30 & 85.49 $\pm$ 8.10 & 91.91 $\pm$ 0.69 \\
\bottomrule
\end{tabular}
\end{adjustbox}
\end{table}

\subsection{\textcolor{black}{Ablation Study}}
\textcolor{black}{We have conducted an ablation study on the IoTID20 dataset to evaluate the contribution of each \textit{Sentinel} component. 
Table~\ref{tab:ablation_iotid20_all} highlights their cumulative impact. Using a standard cross-entropy task loss with Bi-KD establishes a reasonable baseline (\textbf{69.72\%} F1-score at $\alpha=0.1$), but replacing it with a \textbf{class-balanced loss} yields clear gains (\textbf{71.98\%}), confirming its importance for handling skewed attack distributions. Conversely, adding the alignment loss without class balancing provides only modest improvement (\textbf{71.19\%}). This limited effect is expected, since the symmetric teacher and student models naturally produce more congruent feature spaces, reducing the need for strong alignment regularization. The full synergy is realized only when all three components—class-balanced loss, bidirectional KD, and full alignment—are combined in the \textit{Sentinel-I} model, which achieves the top F1-scores of \textbf{77.92\%} ($\alpha=0.1$), \textbf{87.61\%} ($\alpha=1$), and \textbf{92.42\%} (IID). Strikingly, removing the class-balanced loss from this full configuration causes a sharp degradation from \textbf{77.92\%} down to \textbf{71.19\%} on IoTID20, underscoring its role as the critical foundation upon which KD and alignment operate effectively. Overall, these results demonstrate that the modules are not merely additive but mutually reinforcing, with class balancing providing the stable, class-aware foundation required for robust federated intrusion detection.}

\paragraph{\textcolor{black}{Complexity Analysis}} 
\textcolor{black}{We analyze the computational and communication complexity of \text{Sentinel}, comparing it with representative FL methods. Let $R$ denote the number of communication rounds, $E$ the number of local epochs, $N_c$ the number of local data samples, and $B$ the mini-batch size; the effective number of iterations per client per round is $I = E \cdot \tfrac{N_c}{B}$. Parameters $P$ and $Q$ represent the gradient decomposition dimensions in FedKD, and $K$ represents the number of inner-loop proximal updates per round in pFedMe. Among existing approaches, only Sentinel and FedKD adopt a dual-model architecture with distinct teacher ($20{,}741$ parameters) and student ($7{,}813$ parameters) components, whereas baselines such as FedAvg, FedProx, SCAFFOLD, pFedMe, Ditto, FedRep, APPLE, and FedALA rely on single monolithic models.}  

\textcolor{black}{Sentinel (Sentinel-II) achieves a communication complexity of $\mathcal{O}(R \cdot |\theta_S|)$ by transmitting only the lightweight student model, 
like FedKD. However, FedKD employs singular value decomposition (SVD) to further compress the student model, which comes at the cost of performance, particularly degrading minority-class recognition, as evidenced by its lower F1-score in our experiments. In contrast, methods such as APPLE incur substantially higher communication overhead, $\mathcal{O}(RN \cdot |\theta|)$, due to adaptive aggregation across $N$ clients.}  

\textcolor{black}{The computational complexity of Sentinel is}
\[\textcolor{black}{
\mathcal{O}\big(R I (|\theta_T| + |\theta_S| + h_T h_S)\big),}
\]
\textcolor{black}{dominated by teacher training and feature alignment operations. For comparison, the baseline complexities are as follows:}  
\begin{itemize}
    \item \textcolor{black}{Single-model methods (FedAvg, FedProx): $\mathcal{O}(R I \cdot |\theta|)$.} 
    \item \textcolor{black}{Personalized approaches (pFedMe with bi-level optimization): $\mathcal{O}(R (I + K) \cdot |\theta|)$.}  
    \item \textcolor{black}{Higher-order methods (PerFedAvg, 2$^{\text{nd}}$-order MAML): $\mathcal{O}(R I \cdot |\theta|^2)$.}  
    \item \textcolor{black}{FedKD: $\mathcal{O}\!\big(R I (|\theta_T| + |\theta_S| + P Q^{2})\big)$}  
\end{itemize}  

\textcolor{black}{Overall, despite introducing additional local computation, Sentinel preserves low communication overhead by transmitting only the compact student model, making it well-suited for resource-constrained environments that demand efficient and personalized learning.}

\section{Conclusion and Future Scope}
\textcolor{black}{In this paper, we introduced \textit{Sentinel}, an novel personalized federated intrusion detection framework designed to overcome the critical challenges of non-IID data, severe class imbalance, and communication overhead inherent in heterogeneous IoT networks. Our dual-model architecture, which equips each client with a personalized teacher model and a lightweight shared student model, successfully balances deep local adaptation with efficient global knowledge consensus. This design not only preserves client privacy but also significantly reduces communication costs by transmitting only the compact student model.
We have demonstrated that the synergistic integration of three core mechanisms---(1) bidirectional knowledge distillation with adaptive temperature scaling, (2) multi-faceted feature alignment, and (3) class-balanced loss functions---is essential for achieving robust performance. Furthermore, our use of equal-weight normalized gradient aggregation at the server level ensures fairness and enhances stability against client drift. Extensive experiments on the IoTID20 and 5GNIDD datasets validate that \textit{Sentinel} establishes a new state-of-the-art, delivering substantial performance gains over existing federated methods, especially under extreme data heterogeneity where others falter.
For future work, we will explore more advanced personalization, such as prototype-based learning, and develop dynamic aggregation strategies that weigh client updates by quality or relevance to enhance system robustness. Extending \textit{Sentinel} to continual and asynchronous settings is also a crucial next step towards deploying autonomous intrusion detection systems in evolving, real-world environments.}

\bibliographystyle{IEEEtran}
\bibliography{bare_jrnl_new_sample4}
\end{document}